\documentclass[11pt,a4paper]{article}
\usepackage[hyperref]{emnlp-ijcnlp-2019}
\usepackage{times}
\usepackage{latexsym}

\usepackage{url}

\aclfinalcopy 

\usepackage[flushleft]{threeparttable}
\usepackage{todonotes}
\usepackage{enumitem}
\usepackage[linesnumbered,ruled,vlined]{algorithm2e}
\usepackage{algpseudocode}
\usepackage{booktabs}
\usepackage{subcaption}
\usepackage[nointegrals]{wasysym}
\usepackage{tikz}
\usepackage{tabu}
\usepackage{tabularx}
\usepackage{float}
\usepackage{listings}
\usepackage{color}
\usepackage{multirow}
\usepackage{framed}
\usepackage{pifont}
\usepackage{dsfont}
\usepackage{xcolor}
\usepackage{amsmath,amsthm}
\usepackage{amssymb}
\usepackage{graphicx}
\usepackage{textcomp}
\usepackage{tcolorbox}
\usepackage{soul}
\usepackage{todonotes}

\definecolor{DarkRed}{RGB}{130,25,0}

\newcommand{\sentence}[1]{\textit{``#1''}}
\newcommand{\event}[1]{\textsc{{#1}}}
\newcommand{\timex}[1]{\textit{{#1}}}
\newcommand{\temprel}[1]{{\em #1}}
\newcommand{\question}[1]{\textit{``#1''}}

\newcommand{\qiangchange}[1]{{\color{black} {#1}}}

\newcommand{\best}[1]{\textbf{#1}}
\newcommand{\largevar}[1]{\underline{#1}}

\newcommand{\squad}{\textsc{SQuAD}}

\newcommand{\ignore}[1]{}
\definecolor{mypurple}{RGB}{121,55,196}
\definecolor{myblue}{RGB}{60,177,245}
\definecolor{myorange}{RGB}{243,206,84}
\definecolor{mygreen}{RGB}{41,222,35}

\newcounter{exctr}

\newcounter{eventCtr}

\newcounter{timexCtr}

\newcommand{\dataset}{\textsc{Torque}}

\newif\ifcomments
\commentstrue
\ifcomments
    \providecommand{\matt}[1]{{\protect\color{teal}{[Matt: #1]}}}
\else
    \providecommand{\matt}[1]{}
\fi

\newcommand{\qn}[1]{\textcolor{black}{#1}}

\title{
\dataset{}: A Reading Comprehension Dataset of\\ Temporal Ordering Questions
}
\author{Qiang Ning$^\diamondsuit$ ~ Hao Wu$^{\clubsuit}$ ~ Rujun Han$^{\spadesuit}$ ~ Nanyun Peng$^{\spadesuit}$ ~ Matt Gardner$^{\diamondsuit}$ ~ Dan Roth$^{\heartsuit}$\\
$^{\diamondsuit}$Allen Institute for AI \quad $^{\clubsuit}$Hooray Data Co., Ltd \hfill \\
$^{\spadesuit}$University of Southern California \quad $^{\heartsuit}$University of Pennsylvania\\
{\tt \{qiangn,mattg\}@allenai.org},
{\tt haowu@hooray.ai}\\
{\tt \{rujunhan,npeng\}@isi.edu},
{\tt danroth@seas.upenn.edu} }

\date{}

\begin{document}
\maketitle
\begin{abstract}


A critical part of reading is being able to understand the temporal relationships between events described in a passage of text, even when those relationships are not explicitly stated. However, current machine reading comprehension benchmarks have practically no questions that test temporal phenomena, so systems trained on these benchmarks have no capacity to answer questions such as ``what happened \temprel{before/after} [some event]?'' We introduce \dataset{}, a new English reading comprehension benchmark built on 3.2k news snippets with 21k human-generated questions querying temporal relationships. Results show that RoBERTa-large achieves an exact-match score of 51\% on the test set of \dataset{}, about 30\% behind human performance.\footnote{\url{https://allennlp.org/torque.html}}
\end{abstract}
\section{Introduction}
\label{sec:intro}

Time is important for understanding events and stories described in natural language text such as news articles, social media, financial reports, and electronic health records \citep{VGSHKP07,VSCP10,ULADVP13,MSAAEMRUK15,BSCDPV16,BSPP17,LXEBP18}.
For instance, \sentence{he won the championship yesterday} is different from \sentence{he will win the championship tomorrow}: he may be celebrating if he has already won it, while if he has not, he is probably still preparing for the game tomorrow.

The exact time of an event is often implicit in text. 
For instance, if we read that a woman is \sentence{expecting the birth of her first child}, we know that the birth is in the future, while if she is \sentence{mourning the death of her mother}, the death is in the past.
These relationships between an event and a time point (e.g., \sentence{\underline{won} the championship \underline{yesterday}}) or between two events (e.g., ``expecting'' is \temprel{before} ``birth'' and ``mourning'' is \temprel{after} ``death'') are called {\em temporal relations} \cite{PHSSGSRSDFo03}. 

\begin{figure}[t]
    \centering
    \includegraphics[width=0.483\textwidth]{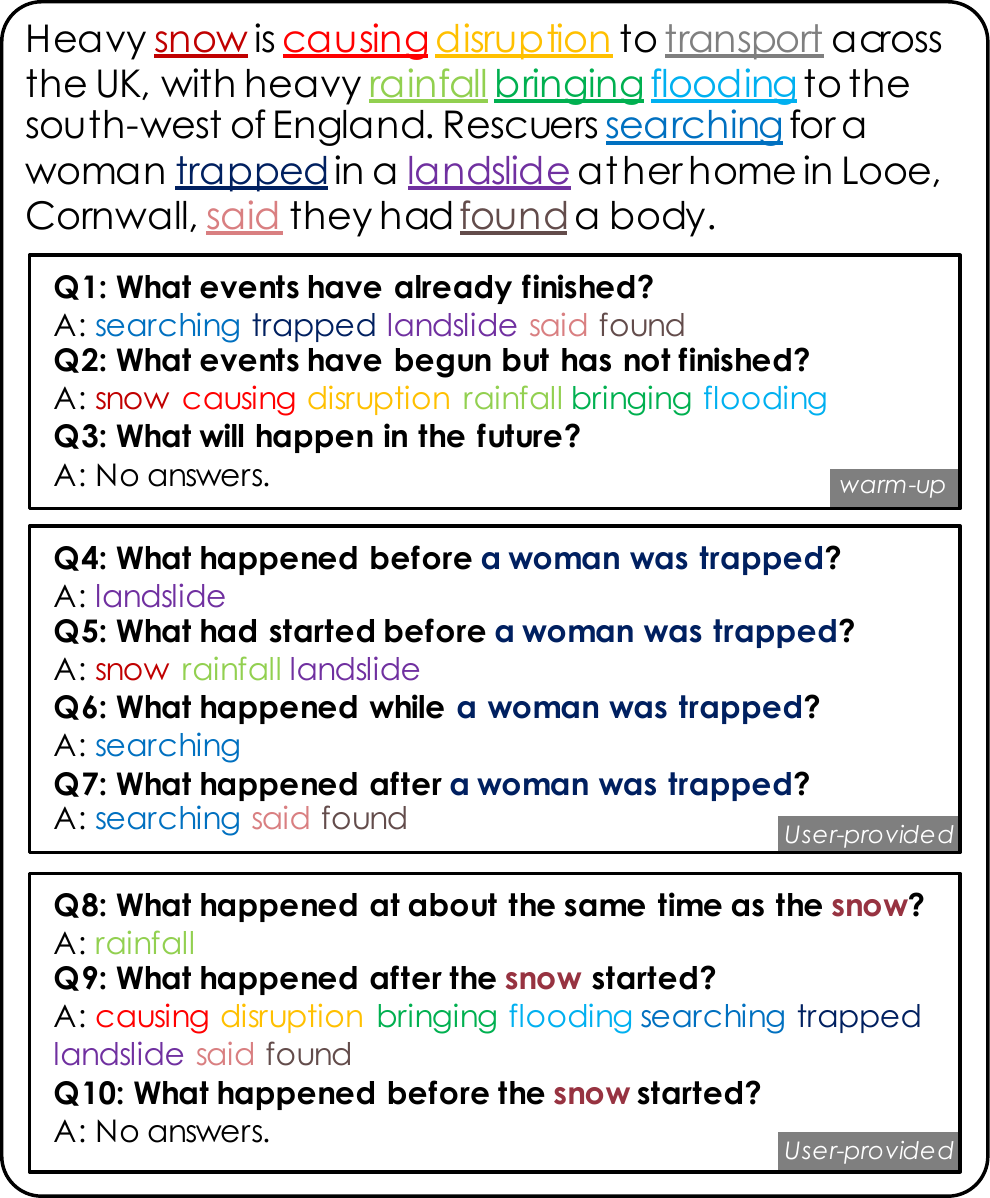}
    \caption{Example annotation of \dataset{}. Events are highlighted in color and contrast questions are grouped.}
    \label{fig:example annotation}
    \vspace{-1em}
\end{figure}

This work studies reading comprehension for temporal relations,
i.e., given a piece of text, a computer needs to answer temporal relation questions (Fig.~\ref{fig:example annotation}).
\qiangchange{Reading comprehension is a natural format for studying temporal phenomena, as the flexibility of natural language annotations allows for capturing relationships that were not possible in previous formalism-based works.}
\qiangchange{However, temporal phenomena are} studied very little in reading comprehension \cite{RZLL16,RajpurkarJiLi18,DWDSSG19,DLMSG19,LTCG19}, and existing systems are hence brittle when handling questions in \dataset{} (Table~\ref{tab:difficulty}).

\begin{figure*}[htbp!]
    \centering
    \includegraphics[width=\textwidth]{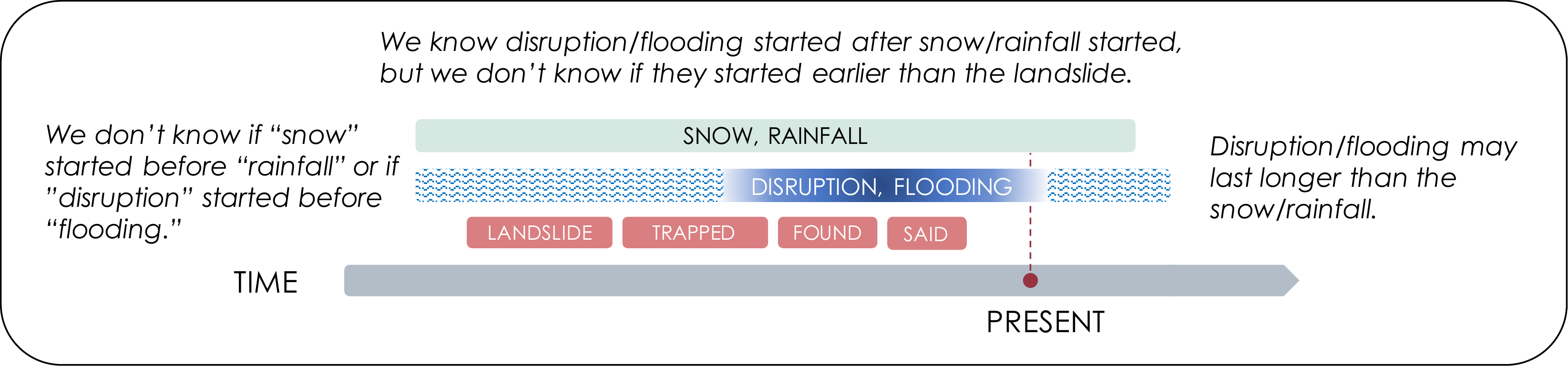}
    \caption{Timeline of the passage in Fig.~\ref{fig:example annotation}.}
    \label{fig:ex timeline}
\end{figure*}

\begin{table*}[htbp!]
    \centering\small
    \begin{tabular}{c|c|c}
        \hline
        Question & BERT (trained on SQuAD) & BERT (trained on SQuAD 2.0)\\
        \hline
        What happened before a woman was trapped? & a landslide & a landslide \\
        What happened after a woman was trapped? & they had found a body & a landslide \\
        What happened while a woman was trapped? & a landslide & a landslide \\
        \hline
        What happened before the snow started? & landslide & heavy rainfall \dots landslide \\
        What happened after the snow started? & flooding to \dots England & heavy rainfall \dots England\\
        What happened during the snow? & a landslide & landslide\\
        \hline
        What happened before the rescuers found a body? & a landslide & a landslide\\
        What happened after the rescuers found a body? & Rescuers searching \dots Cornwall &landslide\\
        What happened during the rescue? & a landslide & they had found a body\\
        \hline
    \end{tabular}
    \begin{tablenotes}
    \scriptsize{
    \item \qiangchange{BERT (SQuAD): \url{https://cogcomp.seas.upenn.edu/page/demo_view/QuASE}}
    \item BERT (SQuAD 2.0): \url{https://www.pragnakalp.com/demos/BERT-NLP-QnA-Demo/}
    }
    \end{tablenotes}
    \caption{\qiangchange{Example system outputs. The correct answers can be seen from the timeline depicted in Fig.~\ref{fig:ex timeline}.}}
    \label{tab:difficulty}
    \vspace{-1em}
\end{table*}

Reading comprehension for temporal relationships has the following challenges.
First, \qiangchange{reading comprehension works rarely require {\em event} understanding.}
For the example in Fig.~\ref{fig:example annotation}, \squad{} \cite{RZLL16} and most datasets largely only require an understanding of predicates and arguments, and would ask questions like \question{what was a woman trapped in?} But a temporal relation question would be \question{what started before a woman was trapped?} To answer it, the system needs to identify events (e.g., \event{landslide} is an event and ``body'' is not), the time of these events (e.g., \event{landslide} is a correct answer, while \event{said} is not because of the time when the two events happen), and look at the entire passage rather than the local predicate-argument structures within a sentence (e.g., \event{snow} and \event{rainfall} are correct answers \qn{to the question above}).

Second, there are many events in a typical passage of text, so temporal relation questions typically query more than one relationship at the same time. This means that a question can have multiple answers (e.g., \question{what happened after the landslide?}), or no answers, because the question may be beyond the time scope (e.g., \question{what happened before the snow started?}).

Third, temporal relations queried by natural language questions are often sensitive to a few key words such as \temprel{before}, \temprel{after}, and \temprel{start}. Those questions can easily be changed to make contrasting questions with dramatically different answers. Models that are not sensitive to these small changes in question words will perform poorly on this task, as shown in Table~\ref{tab:difficulty}.

In this paper, we present \dataset{}, the first reading comprehension benchmark that targets these challenges.
We trained crowd workers to label events in text, and to write and answer questions that query temporal relationships between these events.  We also had workers write questions with contrasting changes to the temporal keywords, to give a comprehensive test of a machine's temporal reasoning ability and minimize the effect of any data collection artifacts \cite{GABBBCDDEGetal20}.
We annotated 3.2k text snippets randomly selected from the TempEval3 dataset \cite{ULADVP13}. In total, \dataset{} has 25k events and 21k user-generated and fully answered temporal relation questions. 20\% of \dataset{} was further validated by additional crowd workers to be used as test data. 
Results show that RoBERTa-large \cite{LOGDJCLLZS19} achieves 51\% 
in exact-match on \dataset{} after fine-tuning, about 30\% behind human \qn{performance}, indicating that more investigation is needed to better solve this problem.

\section{Definitions}
\label{sec:defs}
\subsection{Events}
\label{subsec:event}
As temporal relations are relationships between events, we first define \emph{events}.
Generally speaking, an event involves a predicate and its arguments \cite{ACE05,MYHSBKS15}.
When studying {\em time}, events were defined as actions/states triggered by verbs, adjectives, and nominals \cite{PHSSGSRSDFo03}.
Later works on event and time have largely followed this definition, e.g., TempEval \cite{VGSHKP07}, TimeBank-Dense \cite{CCMB14}, RED \cite{GormanWrPa16}, and MATRES \cite{NingWuRo18}.

This work follows this line of event definition and uses {\em event} and {\em event trigger} interchangeably. 
We define an event to be either a verb or a noun (e.g., \event{trapped} and \event{landslide} in Fig.~\ref{fig:example annotation}).
Specifically, in copular constructions, we choose to label the verb as the event, instead of an adjective or preposition.  This allows us to give a consistent treatment of \sentence{she \underline{was} on the east coast yesterday} and \sentence{she \underline{was} happy}, which we can easily teach to crowd workers.
\qn{Note that from the perspective of data collection, labeling the copula does not lose information as one can always do post-processing using dependency parsing or semantic role labeling to recover the connection between ``was'' and ``happy.''
}

\begin{figure}[t!]
    \centering
    \includegraphics[width=0.48\textwidth]{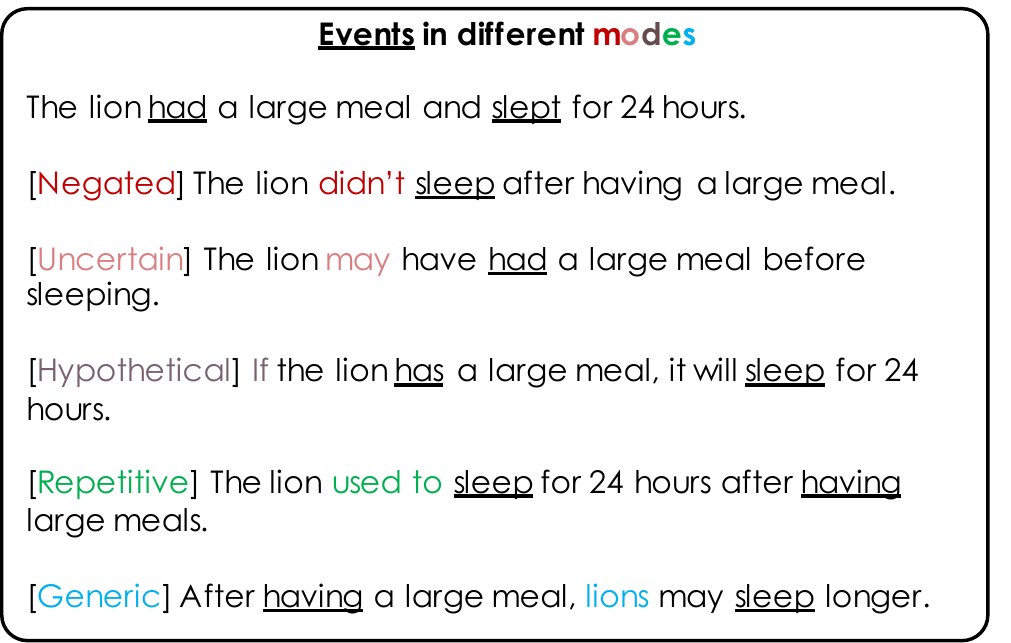}
    \caption{Various modes of events that prior work needed to categorize. Section~\ref{sec: temprel qa} shows that they can be handled naturally without explicit categorization.}
    \label{ex:general events}
\end{figure}

Note that events expressed in text are not always factual.  They can be negated, uncertain, hypothetical, or have other associated modalities (see Fig.~\ref{ex:general events}).  Prior work dealing with events often tried to categorize and label these various aspects because they were crucial for determining temporal relations.  Sometimes certain categories were even dropped due to annotation difficulties \cite{PHSSGSRSDFo03,GormanWrPa16,NingWuRo18}.  In this work, we simply have people label all events, \qn{irrespective of} their modality, and use natural language to describe relations between them, as discussed in Sec.~\ref{sec: temprel qa}.

\subsection{Temporal Relations}
\label{subsec:temporal relations}
Temporal relations describe the relationship between two events with respect to time, or between one event and a fixed time point (e.g., \timex{yesterday}).\footnote{We could also include relationships between two fixed time points (e.g., compare 2011-03-24 with 2011-04-05), but these are mostly trivial, so we do not discuss them further.}
We can use a triplet, $(A, r, B)$, to represent this relationship, where $A$ and $B$ are events or time points, and $r$ is a temporal relation.
For example, the first sentence in Fig.~\ref{ex:general events} expresses a temporal relation (\event{had}, \temprel{happened before}, \event{slept}).

\begin{figure}[t!]
    \centering
    \hspace*{-.2cm}
    \includegraphics[width=0.5\textwidth]{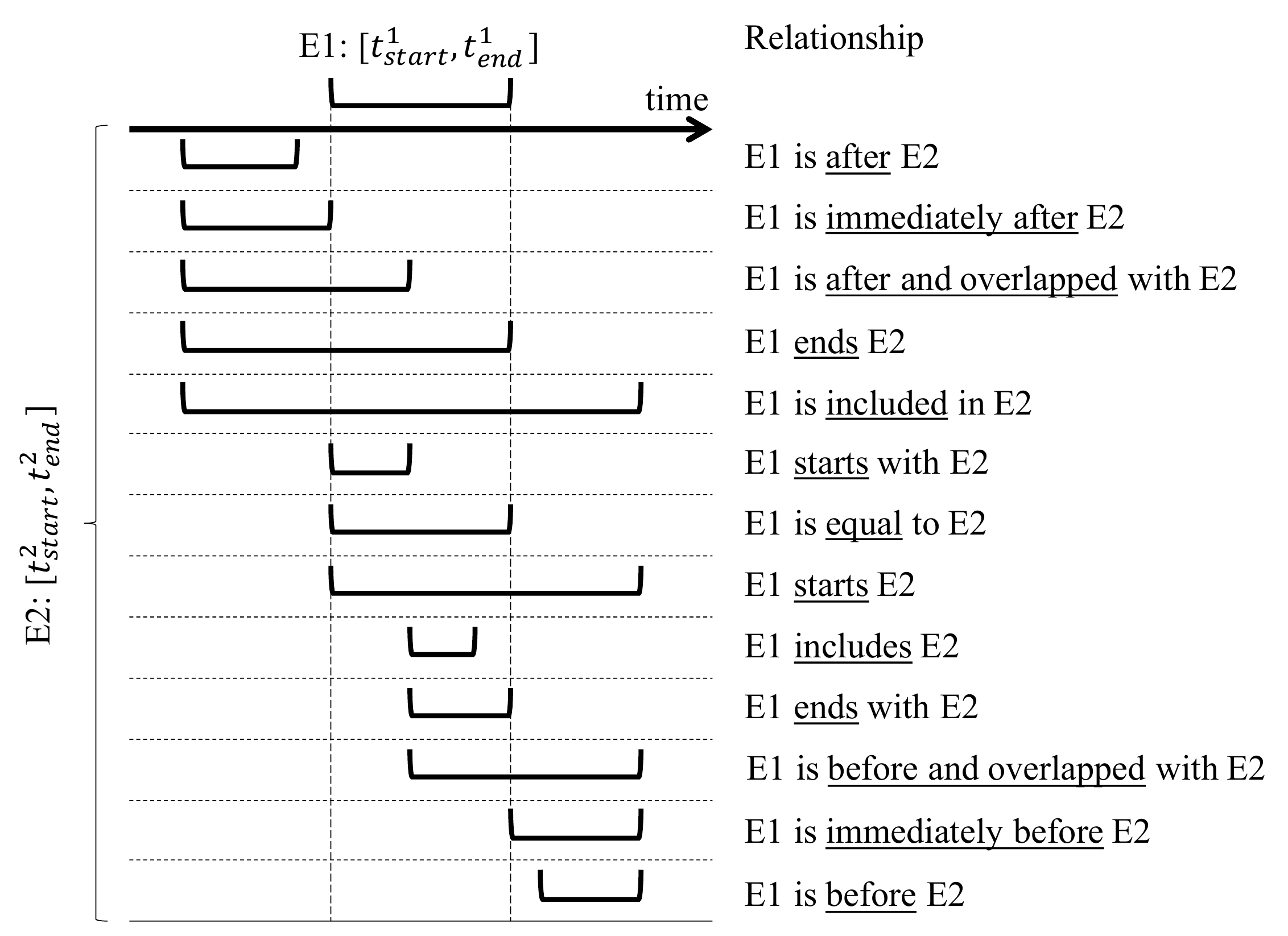}
    \caption{Thirteen relations between two time intervals $[t_{start}^1, t_{end}^1]$ and $[t_{start}^2, t_{end}^2]$. 
    }
    \label{fig:13 labels}
\end{figure}

\begin{figure}[htbp!]
    \centering
    \includegraphics[width=0.49\textwidth]{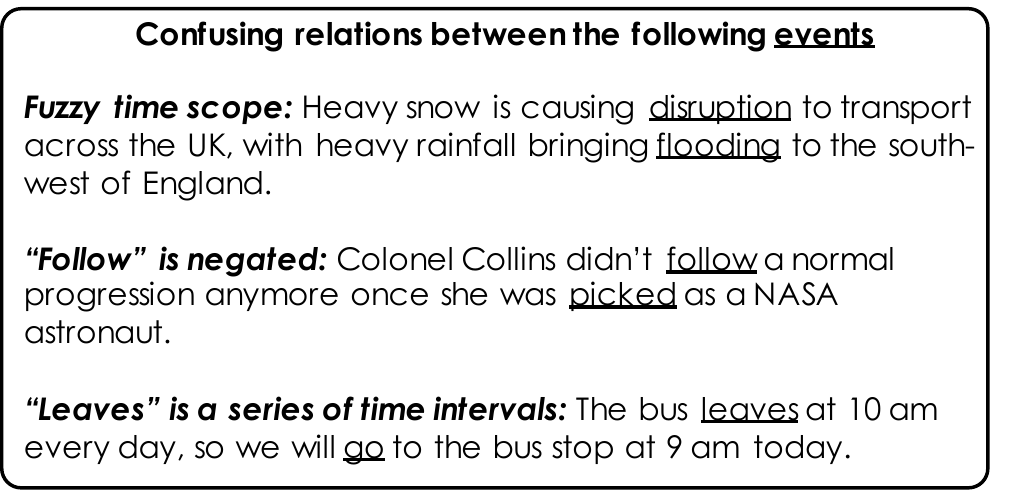}
    \caption{It is confusing to label these relations using a fixed set of relations: they are not simply \temprel{before} or \temprel{after}, but they can be fuzzy, can have modalities as events, and/or need multiple time intervals to represent. 
    }
    \label{fig:confusing relations}
\end{figure}

In previous works, every event is assumed to be associated with a time interval $[t_{start}, t_{end}]$. When comparing two events, there are 13 possible relation labels (see Fig.~\ref{fig:13 labels}) \cite{Allen84}.
However, there are still many relations that cannot be expressed because the assumption that every event has a time interval is inaccurate: The time scope of an event may be fuzzy, an event can have a non-factual modality, or events can be repetitive and invoke multiple intervals (see Fig.~\ref{fig:confusing relations}). To better handle these phenomena, we move away from the fixed set of relations used in prior work and instead use natural language to annotate the relationships between events, as described in the next section.

\section{Natural Language Annotation of Temporal Relations}
\label{sec: temprel qa}

Motivated by recent works \cite{HeLeZe15,MSHDZ17,LSCZ17,Gardner2019QuestionAI}, we propose using natural language question answering as an annotation format for temporal relations.
Recalling that we denote a temporal relation between two events as $(A, r, B)$, we use $(?, r, B)$ to denote a temporal relation question. We instantiate these temporal relation questions using natural language. For instance, (?, \temprel{happened before}, \event{slept}) means \question{what happened before a lion slept?}
We then expect as an answer the set of all events $A$ in the passage such that $(A, r, B)$ holds, assuming for any deictic expression $A$ or $B$ the time point when the passage was written, and assuming that the passage is true.

\begin{figure}[h!]
    \centering
    \includegraphics[width=0.49\textwidth]{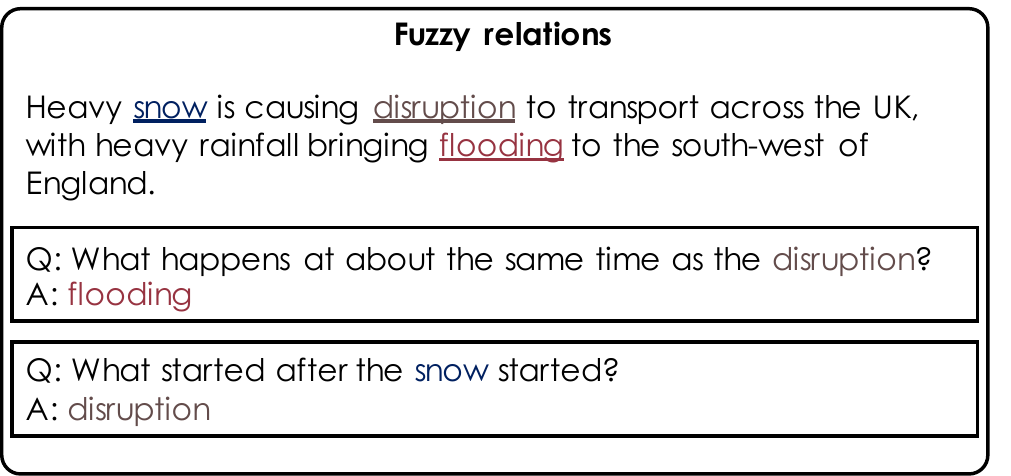}
    \caption{Fuzzy relations that used to be difficult to represent using a predefined label set can be captured naturally in a reading comprehension task. 
    }
    \label{fig:fuzzy relations}
\end{figure}
\subsection{Advantages}
Studying temporal relations as a reading comprehension task gives us the flexibility to handle many of the aforementioned difficulties.
First, fuzzy relations can be described by natural language questions (after all, the relations are expressed in natural language in the first place). In Fig.~\ref{fig:fuzzy relations}, \event{disruption} and \event{flooding} \qn{happened} at about the same time, but we do not know for sure which one is earlier, so we have to choose \temprel{vague}. Similarly for \event{snow} and \event{disruption}, we do not know which one ends earlier and have to choose \temprel{vague} again.
In contrast, the question-answer (QA) pairs in Fig.~\ref{fig:fuzzy relations} can naturally capture these fuzzy relations.


\begin{figure}[t!]
    \centering
    \includegraphics[width=0.49\textwidth]{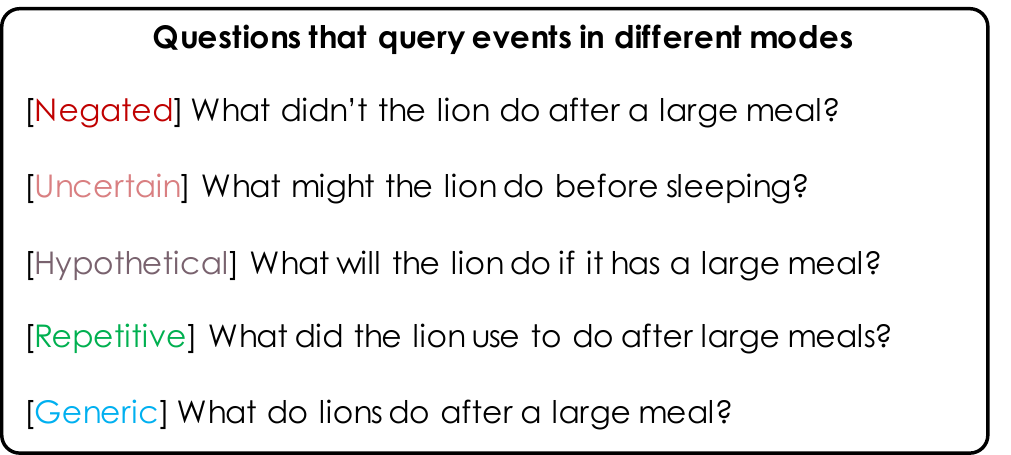}
    \caption{Events in different modes can be distinguished using natural language questions.}
    \label{fig:question modes}
\end{figure}

\begin{figure}[t!]
    \centering
    \includegraphics[width=0.49\textwidth]{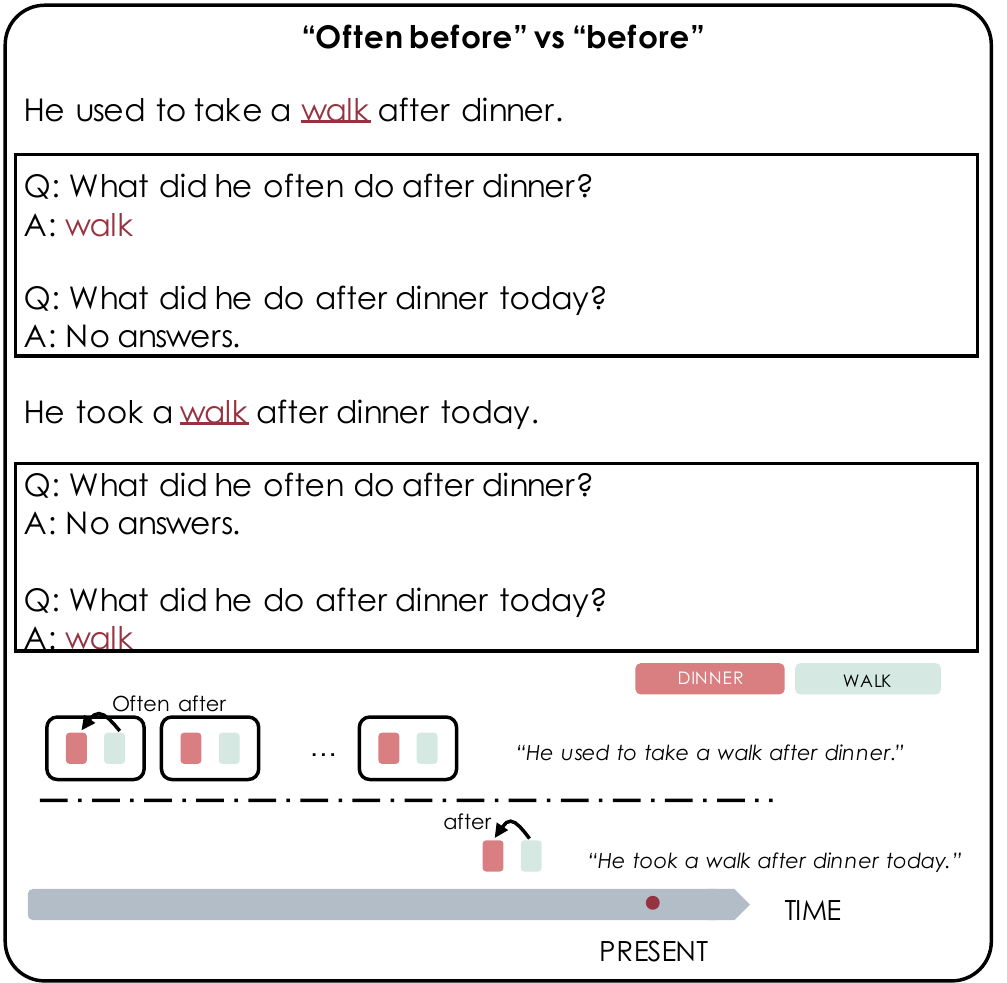}
    \caption{A repetitive event needs multiple time intervals and conveys very different semantics.}
    \label{fig:repetitive}
\end{figure}

Second, natural language questions can conveniently incorporate different modes of events. Figure~\ref{fig:question modes} shows \qiangchange{how to accurately query the relation between \sentence{having a meal} and \sentence{sleeping} in different modes (original sentences can be found in Fig.~\ref{ex:general events}).}
In contrast, if we could only choose one label, we must choose \temprel{before} for all these relations, although these relations are actually different. For instance, a repetitive event may be a series of intervals rather than a single one, and \temprel{often before} is very different from \temprel{before} (Fig.~\ref{fig:repetitive}).



\begin{figure}[t!]
    \centering
    \includegraphics[width=0.49\textwidth]{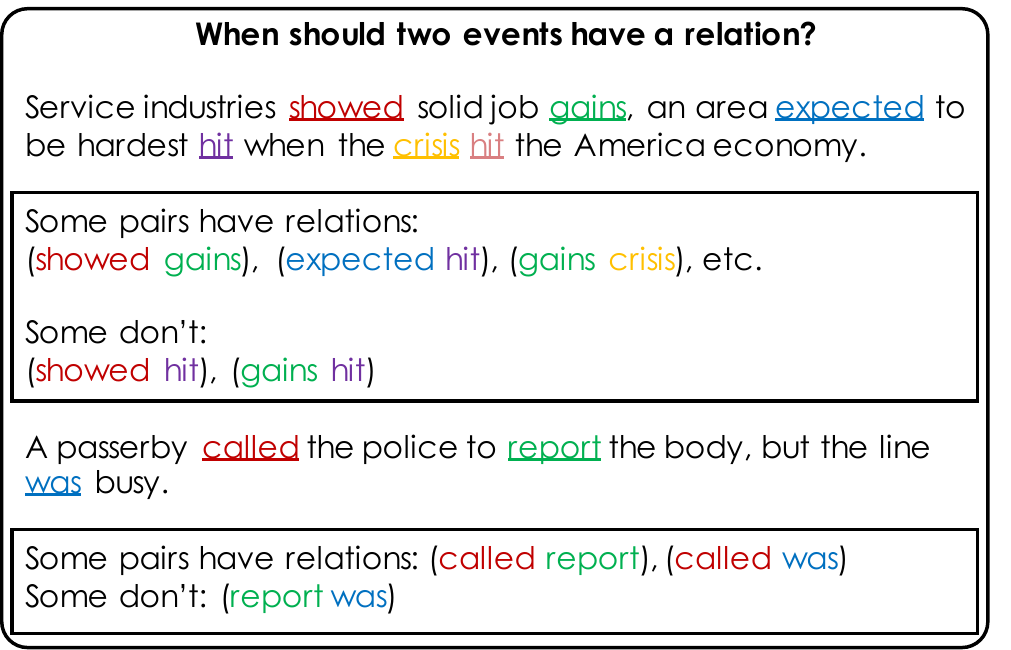}
    \caption{It remains unclear how to determine if two events should have a temporal relation.}
    \label{fig:comparable events}
\end{figure}

Third, a major issue that prior works wanted to address was deciding when two events should have a relation \cite{CMCB14,MGCAV16,GormanWrPa16,NingWuRo18}. 
To avoid asking for relations that do not exist, prior works needed to explicitly annotate certain properties of events as a preprocessing step, but it still remains difficult to have a theory explaining, for instance, why \textcolor{mypurple}{hit} can compare to \textcolor{myblue}{expected} and \textcolor{myorange}{crisis}, but not to \textcolor{mygreen}{gains}.
Interestingly, when we annotate temporal relations in natural language, the annotator naturally avoids event pairs that do not have relations. For instance, for the sentences in Fig.~\ref{fig:comparable events}, one will not ask questions like \question{what happened after the service industries are hardest hit?} or \question{what happened after a passerby reported the body?} Instead, natural questions will be \question{what was expected to happen when the crisis hit America?} and \question{what was supposed to happen after a passerby called the police?} The format of natural language questions bypasses the need for explicit annotation of properties of events or other theories.


\qn{
While using QA as the format gives us many benefits in describing fuzzy relations and incorporating various temporal phenomena, we want to note that it may also lead to potential difficulties in transferring the knowledge to downstream tasks that are not in a QA format, and some special treatment in modeling may be needed (e.g., \citet{HeNiRo19}). This paper focuses on constructing this QA dataset covering new phenomena, and the problem of successful transfer learning is beyond our scope here.
}

\subsection{Penalize Shortcuts by Contrast Sets} 
\begin{figure}[t!]
    \centering
    \includegraphics[width=0.49\textwidth]{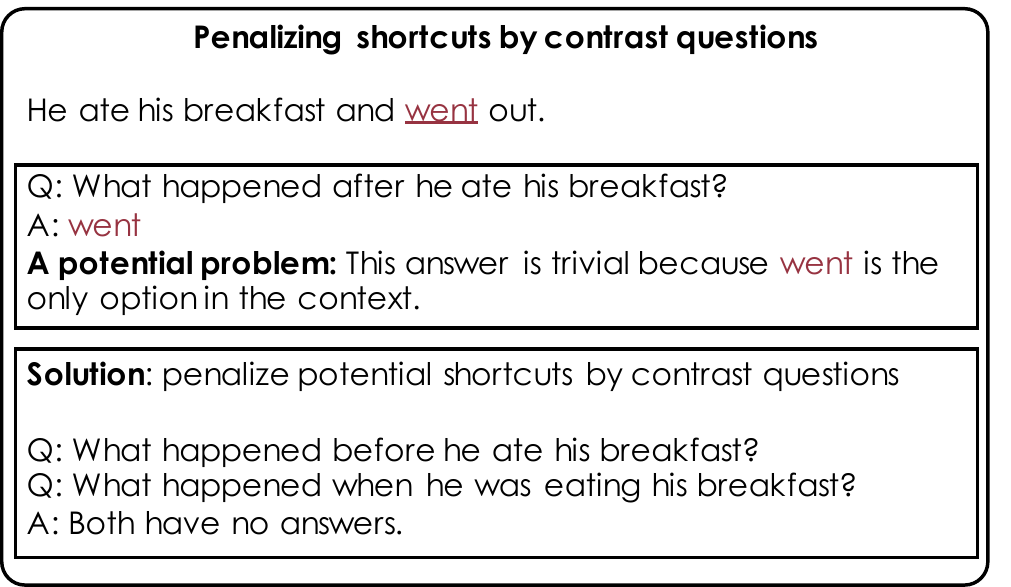}
    \caption{Penalize potential shortcuts by providing contrast questions.}
    \label{fig:short cut}
\end{figure}

\qn{An important problem in building datasets is to avoid trivial solutions \cite{Gardner2019OnMR}.}
As Fig.~\ref{fig:short cut} shows, there are two events \event{ate} and \event{went} in the text. Since \event{ate} is already mentioned in the question, the answer of \event{went} seems a trivial option without the need to understand the underlying relationship.
To address this issue, we create {\em contrast questions} which slightly modify the original questions, but dramatically change the answers, so that shortcuts are penalized. 
Specifically, for an existing question $(?, r, B)$ (e.g., \question{what happened after he ate his breakfast?}), one should keep using $B$ and change $r$ (e.g., \question{what happened \textbf{before}/\textbf{shortly after}/... he ate his breakfast?}), or modify it to ask about the start/end time (e.g., \question{what happened after he \textbf{started} eating his breakfast?} or \question{what would \textbf{finish} after he ate his breakfast?}). We also instructed workers to make sure that the answers to the new question are different from the original one to avoid trivial modifications (e.g., changing \question{what happened} to \question{what occurred}).



\section{Data Collection}
\label{sec:crowdsourcing}
We used Amazon Mechanical Turk to build \dataset{}. 
Following prior work, we focus on passages that consist of two contiguous sentences, as this is sufficient to capture the vast majority of non-trivial temporal relations~\cite{NingFeRo17}. 
We took all the articles used in the TempEval3 (TE3) workshop (2.8k articles) \cite{ULADVP13} and created a pool of 26k two-sentence passages.
Given a random passage from this pool, the annotation process for crowd workers was:

\begin{enumerate}[leftmargin=0.4cm,itemsep=1pt,parsep=0pt]
    \item Label all the events
    \item Repeatedly do the following\footnote{The stopping criterion is discussed in Sec.~\ref{subsec:cost}.}
    \begin{enumerate}[leftmargin=0.6cm,itemsep=0mm]
        \item Ask a temporal relation question and point out all the answers from the list of events
        \item Modify the temporal relation to create one or more new questions and answer them
    \end{enumerate}
\end{enumerate}
\qn{The annotation guidelines\footnote{\scriptsize \url{https://qatmr-qualification.github.io/}} and interface\footnote{\scriptsize\url{https://qatmr.github.io/}} are public.}
In the following sections, we further discuss issues of quality control and crowdsourcing cost.

\subsection{Quality Control}
We used three quality control strategies: qualification, pilot, and validation.

\textbf{Qualification} We designed a separate qualification task where crowd workers were trained and tested on 3 capabilities: labeling events, asking temporal relation questions, and question-answering. 
They were tested on problems randomly selected from a pool we designed. Crowd workers were considered level-1 qualified if they could pass the test within 3 attempts. In practice, about 1 out of 3 workers passed our qualification.

\textbf{Pilot} We then asked level-1 crowd workers to do a small amount of the real task.
We manually checked the annotations and gave feedback to them. Those who passed this inspection were called level-2 workers, and only they could work on the large-scale real task.
Roughly 1 out of 3 pilot submissions received a level-2 qualification.
In the end, there were 63 level-2 annotators, and 60 of them actually worked on our large-scale task.

\textbf{Validation} We randomly selected 20\% of the articles from \dataset{} for further validation. We first validated the events by 4 different level-2 annotators (with the original annotator, there were in total 5 different humans). We also intentionally added noise to the original event list so that the validators must carefully identify wrong events. The final event list was determined by aggregating all 5 humans using majority vote. 
Second, we validated the answers in the same portion of the data. Two level-2 workers were asked to verify the initial annotator's answers; we again added noise to the answer list as a quality control for the validators. Instead of using majority vote as we did for events, the final answers from all workers are considered correct.
We did not do additional validation for the questions themselves, as a manual inspection found the quality to be very high already, with no bad questions in a random sample of 100.

\subsection{Cost}
\label{subsec:cost}
In each job of the main task, we presented 3 passages. The crowd worker could decide to use some or all of them. For each passage a worker decided to use, they needed to label the events, answer 3 hard-coded warm-up questions,
and then ask and answer at least 12 questions (including contrast questions). 
The final reward is a base pay of \$6 plus \$0.5 for each extra question. Crowd workers thus had the incentive to (1) use fewer passages so that they can do event labeling and warm-up questions fewer times, (2) modify questions instead of asking from scratch, and (3) ask extra questions in each job. All these incentives were for more coverage of the temporal phenomena in each passage. In practice, crowd workers on average used 2 passages in each job.
Validating the events in each passage and the answers to a specific question both cost \$0.1.
In total, \dataset{} cost \$15k for an average of \$0.70/question.


\section{\dataset{} Statistics}
\label{sec:stats}
\begin{table*}[t!]
\centering\small
\begin{tabular}{lllr}
\toprule
Type & Subtype & Example & \%\\
\cmidrule(lr){1-1}\cmidrule(lr){2-2}\cmidrule(lr){3-3}\cmidrule(lr){4-4}
Standard & & \question{What happened before Bush gave four key speeches?} & 53\%\\
\cmidrule(lr){1-1}\cmidrule(lr){2-2}\cmidrule(lr){3-3}\cmidrule(lr){4-4}
\multirow{3}{*}{Fuzzy}& begin only & \question{What started before Mr. Fournier was prohibited from organizing his own defense?} & 15\% \\ & overlap only & \question{What events were occurring during the competition?} & 10\% \\ & end only & \question{What will end after he is elected?} & 1\%\\
\cmidrule(lr){1-1}\cmidrule(lr){2-2}\cmidrule(lr){3-3}\cmidrule(lr){4-4}
\multirow{4}{*}{Modality}& uncertain & \question{What might happen after the FTSE 100 index was quoted 9.6 points lower?} & 10\% \\ & negation & \question{What has not taken place before the official figures show something?} & 5\% \\ & hypothetical & \question{What event will happen if the scheme is broadened?} & 2\% \\ & repetitive & \question{What usually happens after common shares are acquired?} & 1\%\\
\cmidrule(lr){1-1}\cmidrule(lr){2-2}\cmidrule(lr){3-3}\cmidrule(lr){4-4}
\multirow{3}{*}{Misc.}& participant & \question{What did Hass do before he went to work as a spy?} & 4\% \\ & opinion & \question{What should happen in the future according to Obama's opinion?} & 3\% \\ & intention & \question{What did Morales want to happen after Washington had a program to eradicate coca?} & 1\%\\
\bottomrule
\end{tabular}
\caption{Temporal phenomena in \dataset{}. ``Standard'' are those that can be directly captured by the previous single-interval-based label set, while other types cannot. Percentages are based on manual inspection of a random sample of 200 questions from \dataset{}; some questions can have multiple types. 
} 
\label{tab:question categorization}
\end{table*}

\dataset{} has 3.2k passage annotations ($\sim$50 tokens/passage),\footnote{Since the passages were selected randomly with replacement, there are 2.9k unique passages in total.} 24.9k events (7.9 events/passage), and 21.2k user-provided questions (half of them were labeled by crowd workers as modifications of existing ones).
Every passage comes with 3 hard-coded warm-up questions asking which events in the passage had already happened, were ongoing, or were still in the future.
Table~\ref{tab:question stats} shows some basic statistics of \dataset{}.
Note the 3 warm-up questions form a contrast set, and we treat the first as ``original'' and the others ``modified.''

\begin{table}[h]
    \centering\small
    \begin{tabular}{lcccc}
    \toprule
                & Q & Q/P & A & A/Q \\
                \midrule
        \textbf{Overall} & 30.7k & 9.7 & 65.0k & 2.1 \\
        Warm-up & 9.5k & 3 & 21.6k & 2.3 \\
        * {\em Original} & 3.2k & 1 & 12.8k & 4.0 \\
        * {\em Modified} & 6.3k & 2 & 8.8k & 1.4 \\
        User-provided & 21.2k & 6.7 & 43.4k & 2.1 \\
        * {\em Original} & 10.6k & 3.4 & 25.1k & 2.4 \\
        * {\em Modified} & 10.6k & 3.3 & 18.3k & 1.7 \\
        \bottomrule
    \end{tabular}
    \caption{Columns from left to right: questions, questions per passage, answers, and answers per question. {\em Modified} is a subset of questions that is created by slightly modifying an {\em original} question.}
    \label{tab:question stats}
\end{table}

In a random sample of 200 questions in the test set of \dataset{}, we found 94 questions querying about relations that cannot be directly represented by the previous single-interval-based labels. Table~\ref{tab:question categorization} gives example questions capturing these phenomena.
More analysis of the event, answer, and workload distributions are in \qiangchange{Appendix~\ref{app:sec:event}-\ref{app:sec:workload}}.

\subsection{Quality} 

To validate the event annotations, we took the events provided by the initial annotator, added noise, and asked different workers to validate. We also trained an auxiliary event detection model using RoBERTa-large and added its predictions as event candidates.
This tells us about the quality of events in \dataset{} in two ways.
First, the Worker Agreement with Aggregate (WAWA) F$_1$ here is 94.2\%; that is, compare the majority-vote with all annotators, and perform micro-average on all instances.
Second, if an event candidate is labeled by both the initial annotator and the model, then almost all of them (99.4\%) are kept by the validators; if neither the initial annotator nor the model labeled a candidate, the candidate is almost surely removed (0.8\%). As validators did not know which ones were noise or not beforehand, this indicates that the validators could identify noise terms reliably. 

Similarly, the WAWA F$_1$ of the answer annotations is 84.7\%, slightly lower than that for events, which is expected because temporal relation QA is intuitively harder. 
Results show that 12.3\% of the randomly added answer candidates were labeled as correct answers by the validators.
We manually inspected 100 questions and found 11.6\% of the added noise terms were correct answers (very close to 12.3\%), indicating that the validators were actually doing a good job in answer validation.
More details of the metrics and the quality of annotations can be found in \qiangchange{Appendix~\ref{app:sec:wawa}}.





\section{Experiment}
\label{sec:exp}
We split \dataset{} into train (80\% of all the questions), dev (5\%), and test (15\%) and these three parts do not have the same articles.
To solve \dataset{} in an end-to-end fashion, the model here takes as input a passage and a question, then looks at every token in the passage and makes a binary classification of whether this token is an answer to the question or not. 
\qn{
Specifically, the model has a one-layered perceptron on top of BERT \cite{DCLT19} or RoBERTa \cite{LOGDJCLLZS19}, and the input to the perceptron layer is the transformers’ output corresponding to the token we’re looking at.
}
We fine-tuned BERT/RoBERTa (both ``base'' and ``large'') on the training set of \dataset{}.
We fixed batch size = 6 (each instance is a tuple of one passage, one question, and all its answers) with gradient accumulation step = 2 in all experiments.
We selected the learning rate (from $(1e^{-5}, 2e^{-5})$), the training epoch (within 10), and the random seed (from 3 arbitrary ones) based on performance on the dev set of \dataset{}.\footnote{\qiangchange{More reproducibility information in Appendix~\ref{app:sec:reproducibility}.}}
%
To compute an estimate of human performance, one author answered 100 questions from the test set \qiangchange{and compared with crowd workers' annotations.}

Both the human performance and system performances are shown in Table~\ref{tab:exp}. 
We report the standard macro F$_1$ and exact-match (EM) metrics in question answering, and also EM consistency, the percentage of contrast question sets for which a model's predictions match exactly to all questions in a group \cite{GABBBCDDEGetal20}.
We see warm-up questions are easier than user-provided ones because warm-up questions focus on easier phenomena of past/ongoing/future events. In addition, RoBERTa-large is expectedly the best system, but still far behind human performance, trailing by about 30\% in EM.

\begin{table*}[htbp!]
\centering\small
\begin{tabular}{lcccccccccccc}
\toprule
& \multicolumn{3}{c}{Dev} & \multicolumn{9}{c}{Test}\\
\cmidrule(lr){2-4}\cmidrule(lr){5-13}
& \multicolumn{3}{c}{}& \multicolumn{3}{c}{Overall}& \multicolumn{3}{c}{Warm-up questions}& \multicolumn{3}{c}{User-provided}\\
\cmidrule(lr){5-7}\cmidrule(lr){8-10}\cmidrule(lr){11-13}
& F$_1$ & EM & C & F$_1$ & EM & C& F$_1$ & EM & C& F$_1$ & EM & C\\
\cmidrule(lr){1-1}\cmidrule(lr){2-2}\cmidrule(lr){3-3}\cmidrule(lr){4-4}\cmidrule(lr){5-5}\cmidrule(lr){6-6}\cmidrule(lr){7-7}\cmidrule(lr){8-8}\cmidrule(lr){9-9}\cmidrule(lr){10-10}\cmidrule(lr){11-11}\cmidrule(lr){12-12}\cmidrule(lr){13-13}
{\em Human} & - & - & - & {\em 95.3} & {\em 84.5} & {\em 82.5} & {\em 95.7} & {\em 89.7} & {\em 90.9} & {\em 95.1} & {\em 82.4} & {\em 79.3} \\
BERT-base & \largevar{67.6} & \largevar{39.6} & \largevar{24.3} & \largevar{67.2} & \largevar{39.8} & \largevar{23.6} & 72.9 & 46.2 & 28.8 & \largevar{64.8} & \largevar{37.1} & \largevar{21.3}\\
BERT-large & 72.8 & 46.0 & 30.7 & 71.9 & 45.9 & 29.1 & 75.0 & 50.1 & \largevar{30.3} & 70.6 & 44.1 & 28.5 \\
RoBERTa-base & 72.2 & 44.5 & 28.7 & 72.6 & 45.7 & 29.9 & 75.4 & 48.8 & 32.3 & 71.4 & 44.4 & 28.8 \\
RoBERTa-large & \best{75.7} & \best{50.4} & \best{36.0} & \best{75.2} & \best{51.1} & \best{\largevar{34.5}} & \best{77.3} & \best{54.3} & \best{36.1} & \best{74.3} & \best{49.8} & \best{\largevar{33.8}} \\
\bottomrule
\end{tabular}
\caption{Human/system performance on the test set of \dataset{}. System performance is averaged from 3 runs; all std. dev. were $\le4\%$ and those in [1\%, 4\%] are \underline{underlined}. C (consistency) is the percentage of contrast groups for which a model's predictions have $F_1\ge 80\%$ for all questions in a group \cite{GABBBCDDEGetal20}.}
\label{tab:exp}
\end{table*}

We further downsampled the training data to test the performance of RoBERTa. We find that with 10\% of the original training data, RoBERTa fails to learn anything meaningful and simply predicts ``not an answer'' for all tokens. With 50\% of the training data, RoBERTa is slightly lower than but already comparable to that of using the entire training set. This means that the learning curve on \dataset{} is already flat and the current size of \dataset{} may not be the bottleneck for its low performance. \qiangchange{Our data and code are public to facilitate more investigations into \dataset{}.\footnote{\url{https://allennlp.org/torque.html}}}


\begin{figure}[htbp!]
    \centering
    \includegraphics[width=.5\textwidth]{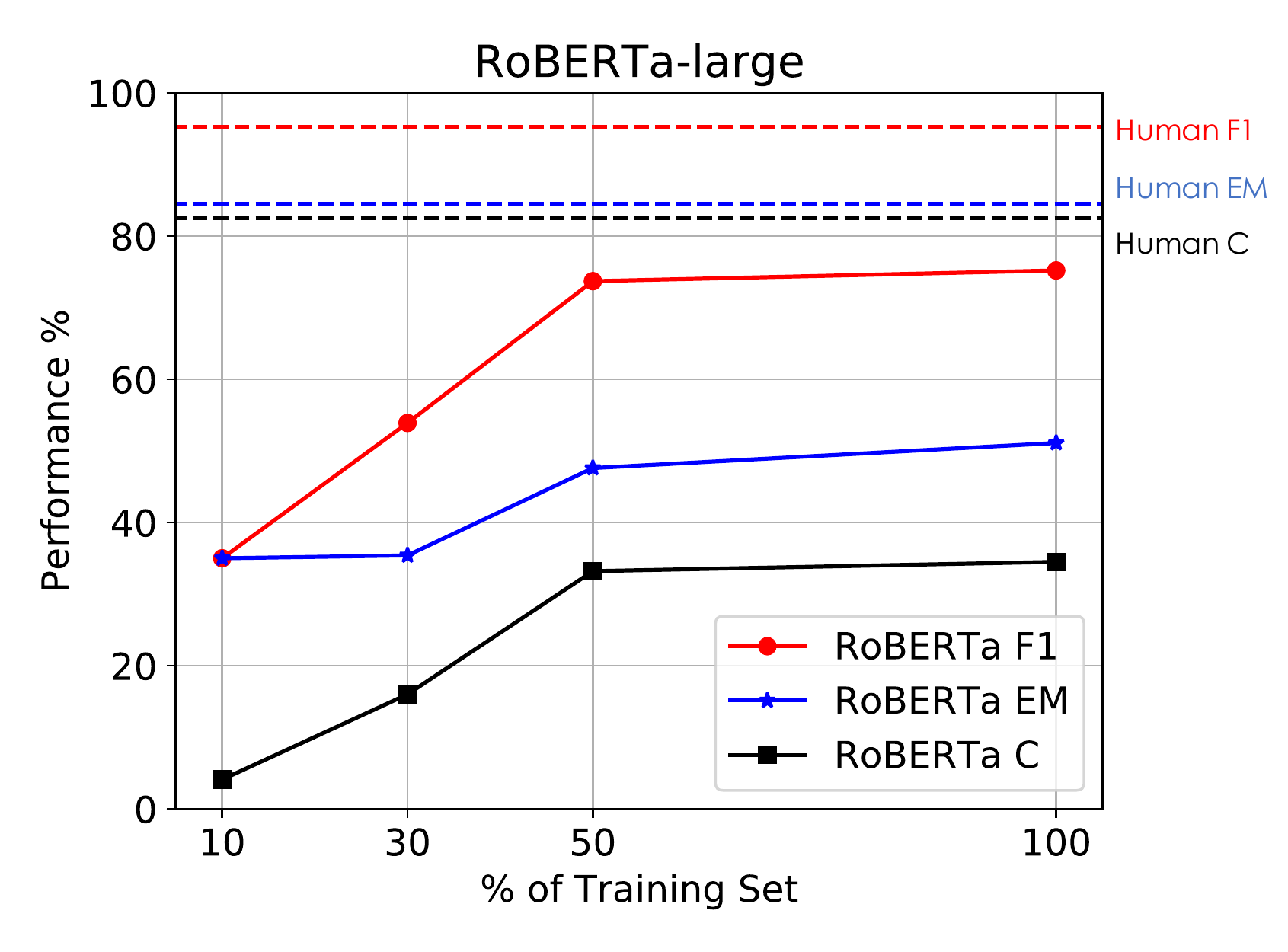}
    \caption{RoBERTa-large with different percentage of training data. Human performance in dashed lines.}
    \label{fig:learning curve}
\end{figure}

\section{Related Work}
\label{sec:related}
The study of {\em time} is to understand {\em when, how long, and how often} things happen.
While {\em how long} and {\em how often} usually require temporal common sense knowledge \cite{VempalaBlPa18,ZKNR19,ZNKR20}, the problem of {\em when} often boils down to extracting temporal relations.

\textbf{Modeling.} Research on temporal relations often focuses on algorithmic improvement, such as structured inference \cite{DoLuRo12,CCMB14,NFWR18}, structured learning \cite{LeeuwenbergMo17,NingFeRo17}, and neural networks \cite{DMLBS17,LMDBS17,TFNT17,ChengMi17,MengRu18,LeeuwenbergMo18,NingSuRo19}.

\textbf{Formalisms.}
The approach that prior works took to handle the aforementioned temporal phenemona was to define formalisms such as the different modes of events (Fig.~\ref{ex:general events}), a predefined label set (Fig.~\ref{fig:13 labels}), different time axes for events \cite{NingWuRo18}, and specific rules to follow when there is confusion.
For example, \citet{BethardMaKl07,NingWuRo18} focused on a limited set of temporal phenomena and achieved high inter-annotator agreements (IAA), while \citet{CMCB14,SBFPPG14,GormanWrPa16} aimed at covering more phenomena but suffered from low IAAs even between NLP researchers.

\textbf{QA as annotation.}
A natural choice is then to cast temporal relation understanding as a machine reading comprehension (MRC) problem.
\dataset{} is motivated by the philosophy in QA-SRL \cite{HeLeZe15} and QAMR \cite{MSHDZ17}, where QA pairs were used as representations for predicate-argument structures. 
In zero-shot relation extraction (RE), they reduced relation slot filling to an MRC problem so as to build very large distant training data and improve zero-shot learning performance \cite{LSCZ17}.
However, our work differs from zero-shot RE since it centers around entities, while \dataset{} is about events; the way to ask and answer questions, \qn{and the way to design a corresponding crowdsourcing pipeline, are} thus significantly different between us.

\qn{The QA-TempEval workshop \cite{LCUMAP15}, desipte its name, is actually not studying temporal relations in an RC setting. The differences between \dataset{} and QA-TempEval are as follows. First, QA TempEval is an evaluation approach for systems that generate TimeML annotations and actually is not a QA task. For instance, QA TempEval is to evaluate whether a system can answer questions like \textsc{``Is $<$Entity\_1$>$ $<$Relation$>$ $<$Entity\_2$>$?''}, where one clearly knows which event that \textsc{$<$Entity$>$} is referring to and where \textsc{Relation} is selected from a predefined label set. Second, QA-TempEval’s annotation relies on the existence of a TimeML corpus. From the perspective of data collection for studying a particular phenomenon, \dataset{} has done more on defining the task and developing a scalable crowdsourcing pipeline. As a result, \dataset{} is also much larger than QA-TempEval and the annotation pipeline of \dataset{} can be easily adopted to collect even more data.
}

\section{Conclusion}
\label{sec:conclusion}
Understanding temporal ordering of events is critical in reading comprehension, but existing works have studied very little about it. This paper presents \dataset{}, a new English machine reading comprehension (MRC) dataset of temporal ordering questions. \dataset{} has 3.2k news snippets, 9.5k hard-coded questions asking which events had happened, were ongoing, or were still in the future, and 21.2k human-generated questions querying more complex phenomena. We argue that an MRC setting allows for more convenient representation of these temporal phenomena than conventional formalisms. Results show that even a state-of-the-art language model, RoBERTa-large, falls behind human performance by a large margin, necessitating more investigation on improving MRC on temporal relationships in the future.


\section*{Acknowledgments}
\qn{
This work was partly supported by contract FA8750-19-2-1004 and contract W911NF-15-1-0543, both with the US Defense Advanced Research Projects Agency (DARPA), and by the Oﬃce of the Director of National Intelligence (ODNI), Intelligence Advanced Research Projects Activity (IARPA), via IARPA Contract No. 2019-19051600006 under the BETTER Program. The views expressed are those of the authors and do not reflect the official policy or position of the Department of Defense or the U.S. Government.
}

\bibliography{ccg-long,cited-long,new}

\begin{thebibliography}{49}
\expandafter\ifx\csname natexlab\endcsname\relax\def\natexlab#1{#1}\fi

\bibitem[{ACE(2005)}]{ACE05}
 2005.
\newblock {The ACE 2005 (ACE 05) Evaluation Plan}.
\newblock Technical report.

\bibitem[{Allen(1984)}]{Allen84}
James~F Allen. 1984.
\newblock Towards a general theory of action and time.
\newblock \emph{Artificial Intelligence}, 23(2):123--154.

\bibitem[{Bethard et~al.(2007)Bethard, Martin, and
  Klingenstein}]{BethardMaKl07}
Steven Bethard, James~H Martin, and Sara Klingenstein. 2007.
\newblock Timelines from text: Identification of syntactic temporal relations.
\newblock In \emph{IEEE International Conference on Semantic Computing (ICSC)},
  pages 11--18.

\bibitem[{Bethard et~al.(2016)Bethard, Savova, Chen, Derczynski, Pustejovsky,
  and Verhagen}]{BSCDPV16}
Steven Bethard, Guergana Savova, Wei-Te Chen, Leon Derczynski, James
  Pustejovsky, and Marc Verhagen. 2016.
\newblock {SemEval}-2016 {Task} 12: {Clinical TempEval}.
\newblock In \emph{Proceedings of the 10th International Workshop on Semantic
  Evaluation (SemEval-2016)}, pages 1052--1062, San Diego, California.
  Association for Computational Linguistics.

\bibitem[{Bethard et~al.(2017)Bethard, Savova, Palmer, and
  Pustejovsky}]{BSPP17}
Steven Bethard, Guergana Savova, Martha Palmer, and James Pustejovsky. 2017.
\newblock Semeval-2017 task 12: Clinical tempeval.
\newblock In \emph{Proceedings of the 11th International Workshop on Semantic
  Evaluation (SemEval-2017)}, pages 565--572. Association for Computational
  Linguistics.

\bibitem[{Cassidy et~al.(2014)Cassidy, McDowell, Chambers, and
  Bethard}]{CMCB14}
Taylor Cassidy, Bill McDowell, Nathanel Chambers, and Steven Bethard. 2014.
\newblock An annotation framework for dense event ordering.
\newblock In \emph{Proceedings of the Annual Meeting of the Association for
  Computational Linguistics (ACL)}, pages 501--506.

\bibitem[{Chambers et~al.(2014)Chambers, Cassidy, McDowell, and
  Bethard}]{CCMB14}
Nathanael Chambers, Taylor Cassidy, Bill McDowell, and Steven Bethard. 2014.
\newblock Dense event ordering with a multi-pass architecture.
\newblock \emph{Transactions of the Association for Computational Linguistics
  (TACL)}, 2:273--284.

\bibitem[{Cheng and Miyao(2017)}]{ChengMi17}
Fei Cheng and Yusuke Miyao. 2017.
\newblock Classifying temporal relations by bidirectional {LSTM} over
  dependency paths.
\newblock In \emph{Proceedings of the Annual Meeting of the Association for
  Computational Linguistics (ACL)}, volume~2, pages 1--6.

\bibitem[{Dasigi et~al.(2019)Dasigi, Liu, Marasovi{\'c}, Smith, and
  Gardner}]{DLMSG19}
Pradeep Dasigi, Nelson~F. Liu, Ana Marasovi{\'c}, Noah~A. Smith, and Matt
  Gardner. 2019.
\newblock {Q}uoref: A reading comprehension dataset with questions requiring
  coreferential reasoning.
\newblock In \emph{Proceedings of the Conference on Empirical Methods in
  Natural Language Processing (EMNLP)}, pages 5925--5932.

\bibitem[{Devlin et~al.(2019)Devlin, Chang, Lee, and Toutanova}]{DCLT19}
Jacob Devlin, Ming-Wei Chang, Kenton Lee, and Kristina Toutanova. 2019.
\newblock {BERT}: Pre-training of deep bidirectional transformers for language
  understanding.
\newblock In \emph{Proceedings of the Conference of the North American Chapter
  of the Association for Computational Linguistics (NAACL)}.

\bibitem[{Dligach et~al.(2017)Dligach, Miller, Lin, Bethard, and
  Savova}]{DMLBS17}
Dmitriy Dligach, Timothy Miller, Chen Lin, Steven Bethard, and Guergana Savova.
  2017.
\newblock Neural temporal relation extraction.
\newblock In \emph{Proceedings of the Conference of the European Chapter of the
  Association for Computational Linguistics (EACL)}, volume~2, pages 746--751.

\bibitem[{Do et~al.(2012)Do, Lu, and Roth}]{DoLuRo12}
Quang Do, Wei Lu, and Dan Roth. 2012.
\newblock Joint inference for event timeline construction.
\newblock In \emph{Proceedings of the Conference on Empirical Methods in
  Natural Language Processing (EMNLP)}.

\bibitem[{Dua et~al.(2019)Dua, Wang, Dasigi, Stanovsky, Singh, and
  Gardner}]{DWDSSG19}
Dheeru Dua, Yizhong Wang, Pradeep Dasigi, Gabriel Stanovsky, Sameer Singh, and
  Matt Gardner. 2019.
\newblock {DROP}: A reading comprehension benchmark requiring discrete
  reasoning over paragraphs.
\newblock In \emph{Proceedings of the Conference of the North American Chapter
  of the Association for Computational Linguistics (NAACL)}.

\bibitem[{Gardner et~al.(2020)Gardner, Artzi, Basmova, Berant, Bogin, Chen,
  Dasigi, Dua, Elazar, Gottumukkala, Gupta, Hajishirzi, Ilharco, Khashabi, Lin,
  Liu, Liu, Mulcaire, Ning, Singh, Smith, Subramanian, Tsarfaty, Wallace,
  Zhang, and Zhou}]{GABBBCDDEGetal20}
Matt Gardner, Yoav Artzi, Victoria Basmova, Jonathan Berant, Ben Bogin, Sihao
  Chen, Pradeep Dasigi, Dheeru Dua, Yanai Elazar, Ananth Gottumukkala, Nitish
  Gupta, Hanna Hajishirzi, Gabriel Ilharco, Daniel Khashabi, Kevin Lin,
  Jiangming Liu, Nelson~F. Liu, Phoebe Mulcaire, Qiang Ning, Sameer Singh,
  Noah~A. Smith, Sanjay Subramanian, Reut Tsarfaty, Eric Wallace, A.~Zhang, and
  Ben Zhou. 2020.
\newblock Evaluating models' local decision boundaries via contrast sets.
\newblock In \emph{Findings of EMNLP}.

\bibitem[{Gardner et~al.(2019{\natexlab{a}})Gardner, Berant, Hajishirzi,
  Talmor, and Min}]{Gardner2019OnMR}
Matt Gardner, Jonathan Berant, Hannaneh Hajishirzi, Alon Talmor, and Sewon Min.
  2019{\natexlab{a}}.
\newblock On making reading comprehension more comprehensive.
\newblock In \emph{Proceedings of the 2nd Workshop on Machine Reading for
  Question Answering}.

\bibitem[{Gardner et~al.(2019{\natexlab{b}})Gardner, Berant, Hajishirzi,
  Talmor, and Min}]{Gardner2019QuestionAI}
Matt Gardner, Jonathan Berant, Hannaneh Hajishirzi, Alon Talmor, and Sewon Min.
  2019{\natexlab{b}}.
\newblock Question answering is a format; when is it useful?
\newblock \emph{arXiv preprint arXiv:1909.11291}.

\bibitem[{Geva et~al.(2019)Geva, Goldberg, and Berant}]{GevaGoBe19}
Mor Geva, Yoav Goldberg, and Jonathan Berant. 2019.
\newblock Are we modeling the task or the annotator? an investigation of
  annotator bias in natural language understanding datasets.
\newblock In \emph{Proceedings of the Conference on Empirical Methods in
  Natural Language Processing (EMNLP)}.

\bibitem[{He et~al.(2020)He, Ning, and Roth}]{HeNiRo19}
Hangfeng He, Qiang Ning, and Dan Roth. 2020.
\newblock {Q}u{ASE}: Question-answer driven sentence encoding.
\newblock In \emph{Proceedings of the Annual Meeting of the Association for
  Computational Linguistics (ACL)}, pages 8743--8758.

\bibitem[{He et~al.(2015)He, Lewis, and Zettlemoyer}]{HeLeZe15}
Luheng He, Mike Lewis, and Luke Zettlemoyer. 2015.
\newblock Question-answer driven semantic role labeling: Using natural language
  to annotate natural language.
\newblock In \emph{Proceedings of the Conference on Empirical Methods in
  Natural Language Processing (EMNLP)}, pages 643--653.

\bibitem[{Laparra et~al.(2018)Laparra, Xu, Elsayed, Bethard, and
  Palmer}]{LXEBP18}
Egoitz Laparra, Dongfang Xu, Ahmed Elsayed, Steven Bethard, and Martha Palmer.
  2018.
\newblock {SemEval} 2018 task 6: Parsing time normalizations.
\newblock In \emph{Proceedings of The 12th International Workshop on Semantic
  Evaluation}, pages 88--96.

\bibitem[{Leeuwenberg and Moens(2017)}]{LeeuwenbergMo17}
Artuur Leeuwenberg and Marie-Francine Moens. 2017.
\newblock Structured learning for temporal relation extraction from clinical
  records.
\newblock In \emph{Proceedings of the 15th Conference of the European Chapter
  of the Association for Computational Linguistics}.

\bibitem[{Leeuwenberg and Moens(2018)}]{LeeuwenbergMo18}
Artuur Leeuwenberg and Marie-Francine Moens. 2018.
\newblock Temporal information extraction by predicting relative time-lines.
\newblock \emph{Proceedings of the Conference on Empirical Methods in Natural
  Language Processing (EMNLP)}.

\bibitem[{Levy et~al.(2017)Levy, Seo, Choi, and Zettlemoyer}]{LSCZ17}
Omer Levy, Minjoon Seo, Eunsol Choi, and Luke Zettlemoyer. 2017.
\newblock Zero-shot relation extraction via reading comprehension.
\newblock In \emph{Proceedings of the SIGNLL Conference on Computational
  Natural Language Learning (CoNLL)}, pages 333--342.

\bibitem[{Lin et~al.(2017)Lin, Miller, Dligach, Bethard, and Savova}]{LMDBS17}
Chen Lin, Timothy Miller, Dmitriy Dligach, Steven Bethard, and Guergana Savova.
  2017.
\newblock Representations of time expressions for temporal relation extraction
  with convolutional neural networks.
\newblock \emph{BioNLP 2017}, pages 322--327.

\bibitem[{Lin et~al.(2019)Lin, Tafjord, Clark, and Gardner}]{LTCG19}
Kevin Lin, Oyvind Tafjord, Peter Clark, and Matt Gardner. 2019.
\newblock Reasoning over paragraph effects in situations.
\newblock In \emph{Proceedings of the 2nd Workshop on Machine Reading for
  Question Answering}, pages 58--62.

\bibitem[{Liu et~al.(2019)Liu, Ott, Goyal, Du, Joshi, Chen, Levy, Lewis,
  Zettlemoyer, and Stoyanov}]{LOGDJCLLZS19}
Yinhan Liu, Myle Ott, Naman Goyal, Jingfei Du, Mandar Joshi, Danqi Chen, Omer
  Levy, Mike Lewis, Luke Zettlemoyer, and Veselin Stoyanov. 2019.
\newblock {RoBERTa}: A robustly optimized {BERT} pretraining approach.
\newblock \emph{arXiv preprint arXiv:1907.11692}.

\bibitem[{Llorens et~al.(2015)Llorens, Chambers, UzZaman, Mostafazadeh, Allen,
  and Pustejovsky}]{LCUMAP15}
Hector Llorens, Nathanael Chambers, Naushad UzZaman, Nasrin Mostafazadeh, James
  Allen, and James Pustejovsky. 2015.
\newblock {SemEval}-2015 {Task} 5: {QA} {TEMPEVAL} - evaluating temporal
  information understanding with question answering.
\newblock In \emph{Proceedings of the 9th International Workshop on Semantic
  Evaluation (SemEval 2015)}, pages 792--800.

\bibitem[{Meng and Rumshisky(2018)}]{MengRu18}
Yuanliang Meng and Anna Rumshisky. 2018.
\newblock Context-aware neural model for temporal information extraction.
\newblock In \emph{Proceedings of the Annual Meeting of the Association for
  Computational Linguistics (ACL)}, volume~1, pages 527--536.

\bibitem[{Michael et~al.(2017)Michael, Stanovsky, He, Dagan, and
  Zettlemoyer}]{MSHDZ17}
Julian Michael, Gabriel Stanovsky, Luheng He, Ido Dagan, and Luke Zettlemoyer.
  2017.
\newblock Crowdsourcing question-answer meaning representations.
\newblock \emph{arXiv preprint arXiv:1711.05885}.

\bibitem[{Minard et~al.(2015)Minard, Speranza, Agirre, Aldabe, van Erp,
  Magnini, Rigau, Urizar, and Kessler}]{MSAAEMRUK15}
Anne-Lyse Minard, Manuela Speranza, Eneko Agirre, Itziar Aldabe, Marieke van
  Erp, Bernardo Magnini, German Rigau, Ruben Urizar, and Fondazione~Bruno
  Kessler. 2015.
\newblock {SemEval}-2015 {Task} 4: {TimeLine}: Cross-document event ordering.
\newblock In \emph{Proceedings of the 9th International Workshop on Semantic
  Evaluation (SemEval 2015)}, pages 778--786.

\bibitem[{Mitamura et~al.(2015)Mitamura, Yamakawa, Holm, Song, Bies, Kulick,
  and Strassel}]{MYHSBKS15}
T.~Mitamura, Y.~Yamakawa, S.~Holm, Z.~Song, A.~Bies, S.~Kulick, and
  S.~Strassel. 2015.
\newblock Event nugget annotation: Processes and issues.
\newblock In \emph{Proceedings of the Workshop on Events at NAACL-HLT}.

\bibitem[{Mostafazadeh et~al.(2016)Mostafazadeh, Grealish, Chambers, Allen, and
  Vanderwende}]{MGCAV16}
Nasrin Mostafazadeh, Alyson Grealish, Nathanael Chambers, James Allen, and Lucy
  Vanderwende. 2016.
\newblock {CaTeRS}: Causal and temporal relation scheme for semantic annotation
  of event structures.
\newblock In \emph{Proceedings of the 4th Workshop on Events: Definition,
  Detection, Coreference, and Representation}, pages 51--61.

\bibitem[{Ning et~al.(2017)Ning, Feng, and Roth}]{NingFeRo17}
Qiang Ning, Zhili Feng, and Dan Roth. 2017.
\newblock A structured learning approach to temporal relation extraction.
\newblock In \emph{Proceedings of the Conference on Empirical Methods in
  Natural Language Processing (EMNLP)}, pages 1038--1048, Copenhagen, Denmark.
  Association for Computational Linguistics.

\bibitem[{Ning et~al.(2018{\natexlab{a}})Ning, Feng, Wu, and Roth}]{NFWR18}
Qiang Ning, Zhili Feng, Hao Wu, and Dan Roth. 2018{\natexlab{a}}.
\newblock Joint reasoning for temporal and causal relations.
\newblock In \emph{Proceedings of the Annual Meeting of the Association for
  Computational Linguistics (ACL)}, pages 2278--2288. Association for
  Computational Linguistics.

\bibitem[{Ning et~al.(2019)Ning, Subramanian, and Roth}]{NingSuRo19}
Qiang Ning, Sanjay Subramanian, and Dan Roth. 2019.
\newblock {An Improved Neural Baseline for Temporal Relation Extraction}.
\newblock In \emph{Proceedings of the Conference on Empirical Methods in
  Natural Language Processing (EMNLP)}.

\bibitem[{Ning et~al.(2018{\natexlab{b}})Ning, Wu, and Roth}]{NingWuRo18}
Qiang Ning, Hao Wu, and Dan Roth. 2018{\natexlab{b}}.
\newblock A multi-axis annotation scheme for event temporal relations.
\newblock In \emph{Proceedings of the Annual Meeting of the Association for
  Computational Linguistics (ACL)}, pages 1318--1328. Association for
  Computational Linguistics.

\bibitem[{O'Gorman et~al.(2016)O'Gorman, Wright-Bettner, and
  Palmer}]{GormanWrPa16}
Tim O'Gorman, Kristin Wright-Bettner, and Martha Palmer. 2016.
\newblock {Richer Event Description}: Integrating event coreference with
  temporal, causal and bridging annotation.
\newblock In \emph{Proceedings of the 2nd Workshop on Computing News Storylines
  (CNS 2016)}, pages 47--56, Austin, Texas. Association for Computational
  Linguistics.

\bibitem[{Pustejovsky et~al.(2003)Pustejovsky, Hanks, Sauri, See, Gaizauskas,
  Setzer, Radev, Sundheim, Day, Ferro et~al.}]{PHSSGSRSDFo03}
James Pustejovsky, Patrick Hanks, Roser Sauri, Andrew See, Robert Gaizauskas,
  Andrea Setzer, Dragomir Radev, Beth Sundheim, David Day, Lisa Ferro, et~al.
  2003.
\newblock The {TIMEBANK} corpus.
\newblock In \emph{Corpus Linguistics}, page~40.

\bibitem[{Rajpurkar et~al.(2018)Rajpurkar, Jia, and Liang}]{RajpurkarJiLi18}
Pranav Rajpurkar, Robin Jia, and Percy Liang. 2018.
\newblock Know what you don{'}t know: Unanswerable questions for {SQ}u{AD}.
\newblock In \emph{Proceedings of the Annual Meeting of the Association for
  Computational Linguistics (ACL)}, pages 784--789.

\bibitem[{Rajpurkar et~al.(2016)Rajpurkar, Zhang, Lopyrev, and Liang}]{RZLL16}
Pranav Rajpurkar, Jian Zhang, Konstantin Lopyrev, and Percy Liang. 2016.
\newblock {SQ}u{AD}: 100,000+ questions for machine comprehension of text.
\newblock In \emph{Proceedings of the Conference on Empirical Methods in
  Natural Language Processing (EMNLP)}, pages 2383--2392.

\bibitem[{Semega et~al.(2019)Semega, Kollar, Creamer, and Mohanty}]{SKCM18}
Jessica Semega, Melissa Kollar, John Creamer, and Abinash Mohanty. 2019.
\newblock \href
  {https://www.census.gov/content/dam/Census/library/publications/2019/demo/p60-266.pdf}
  {\emph{Income and Poverty in the United States: 2018}}.
\newblock U.S. Department of Commerce.

\bibitem[{Styler~IV et~al.(2014)Styler~IV, Bethard, Finan, Palmer, Pradhan,
  de~Groen, Erickson, Miller, Lin, Savova et~al.}]{SBFPPG14}
William~F Styler~IV, Steven Bethard, Sean Finan, Martha Palmer, Sameer Pradhan,
  Piet~C de~Groen, Brad Erickson, Timothy Miller, Chen Lin, Guergana Savova,
  et~al. 2014.
\newblock Temporal annotation in the clinical domain.
\newblock \emph{Transactions of the Association for Computational Linguistics
  (TACL)}, 2:143.

\bibitem[{Tourille et~al.(2017)Tourille, Ferret, Neveol, and Tannier}]{TFNT17}
Julien Tourille, Olivier Ferret, Aurelie Neveol, and Xavier Tannier. 2017.
\newblock Neural architecture for temporal relation extraction: A {Bi-LSTM}
  approach for detecting narrative containers.
\newblock In \emph{Proceedings of the Annual Meeting of the Association for
  Computational Linguistics (ACL)}, volume~2, pages 224--230.

\bibitem[{UzZaman et~al.(2013)UzZaman, Llorens, Allen, Derczynski, Verhagen,
  and Pustejovsky}]{ULADVP13}
Naushad UzZaman, Hector Llorens, James Allen, Leon Derczynski, Marc Verhagen,
  and James Pustejovsky. 2013.
\newblock {SemEval}-2013 {Task} 1: {TEMPEVAL}-3: Evaluating time expressions,
  events, and temporal relations.
\newblock \emph{Proceedings of the Joint Conference on Lexical and
  Computational Semantics (*SEM)}, 2:1--9.

\bibitem[{Vempala et~al.(2018)Vempala, Blanco, and Palmer}]{VempalaBlPa18}
Alakananda Vempala, Eduardo Blanco, and Alexis Palmer. 2018.
\newblock Determining event durations: Models and error analysis.
\newblock In \emph{Proceedings of the Conference of the North American Chapter
  of the Association for Computational Linguistics (NAACL)}, volume~2, pages
  164--168.

\bibitem[{Verhagen et~al.(2007)Verhagen, Gaizauskas, Schilder, Hepple, Katz,
  and Pustejovsky}]{VGSHKP07}
Marc Verhagen, Robert Gaizauskas, Frank Schilder, Mark Hepple, Graham Katz, and
  James Pustejovsky. 2007.
\newblock {SemEval}-2007 {Task} 15: {TempEval} temporal relation
  identification.
\newblock In \emph{Proceedings of the 4th International Workshop on Semantic
  Evaluations}, pages 75--80. Association for Computational Linguistics.

\bibitem[{Verhagen et~al.(2010)Verhagen, Sauri, Caselli, and
  Pustejovsky}]{VSCP10}
Marc Verhagen, Roser Sauri, Tommaso Caselli, and James Pustejovsky. 2010.
\newblock {SemEval}-2010 {Task} 13: {TempEval}-2.
\newblock In \emph{Proceedings of the 5th International Workshop on Semantic
  Evaluation}, pages 57--62. Association for Computational Linguistics.

\bibitem[{Zhou et~al.(2019)Zhou, Khashabi, Ning, and Roth}]{ZKNR19}
Ben Zhou, Daniel Khashabi, Qiang Ning, and Dan Roth. 2019.
\newblock {``Going on a vacation'' takes longer than ``Going for a walk'': A
  Study of Temporal Commonsense Understanding}.
\newblock In \emph{Proceedings of the Conference on Empirical Methods in
  Natural Language Processing (EMNLP)}.

\bibitem[{Zhou et~al.(2020)Zhou, Ning, Khashabi, and Roth}]{ZNKR20}
Ben Zhou, Qiang Ning, Daniel Khashabi, and Dan Roth. 2020.
\newblock {Temporal Common Sense Acquisition with Minimal Supervision}.
\newblock In \emph{Proceedings of the Annual Meeting of the Association for
  Computational Linguistics (ACL)}.

\end{thebibliography}
\bibliographystyle{acl_natbib}
\newpage
\onecolumn
\appendix

\section*{Appendix}
\section{Event Distribution}
\label{app:sec:event}
\begin{figure}[h!]
    \centering
    \includegraphics[width=.5\textwidth]{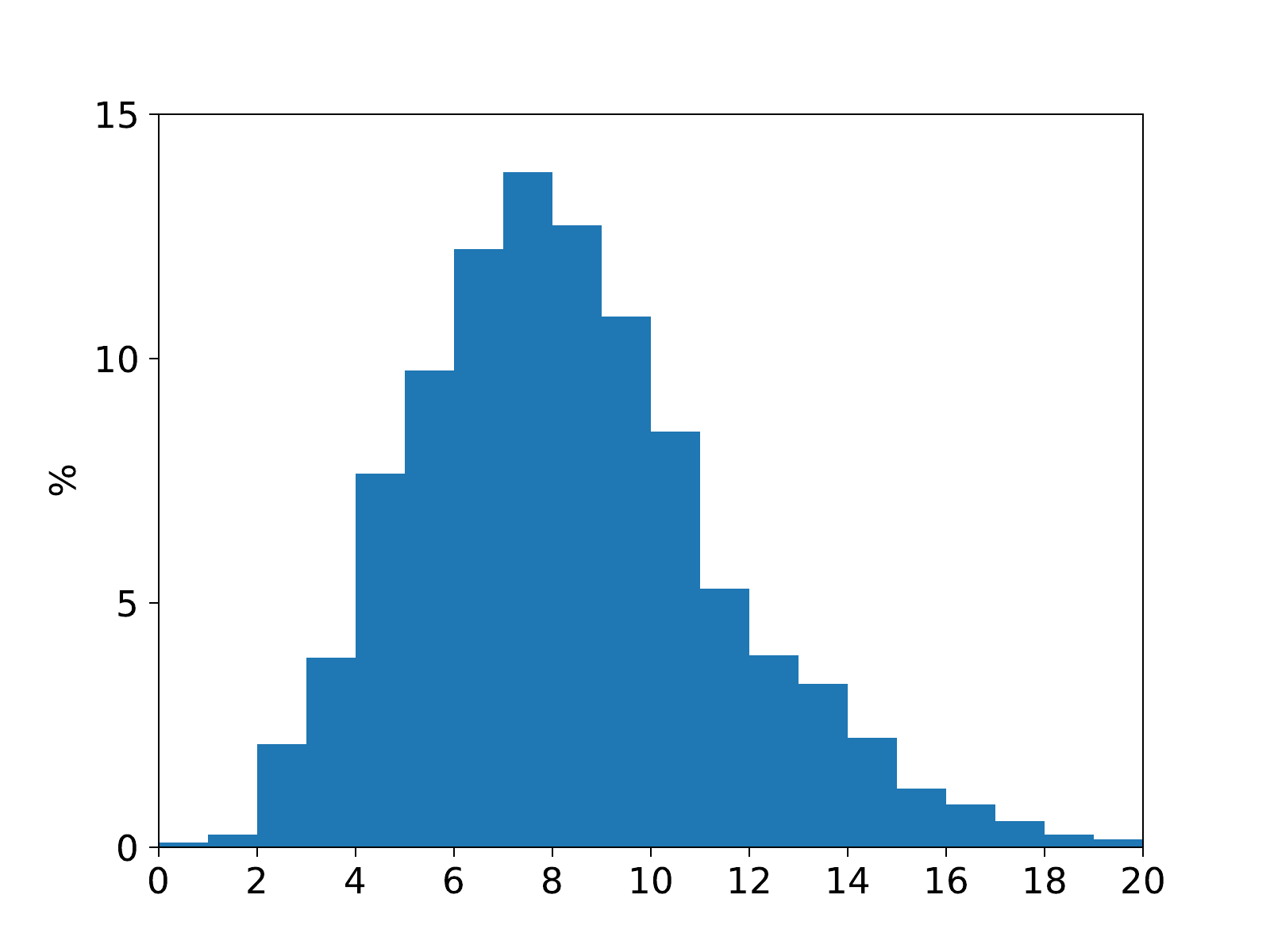}
    \caption{Histogram of the number of events in all passages in \dataset{}.}
    \label{fig:event num hist}
\end{figure}

As we mentioned in Sec.~\ref{sec:stats}, \dataset{} has 24.9k events over 3.2k passages. Figure~\ref{fig:event num hist} shows the histogram of the number of events in all these passages. We can see it roughly follows a Gaussian distribution with the mean at around 7-8 events per passage.

\begin{figure*}[h!]
    \centering
    \includegraphics[width=.8\textwidth]{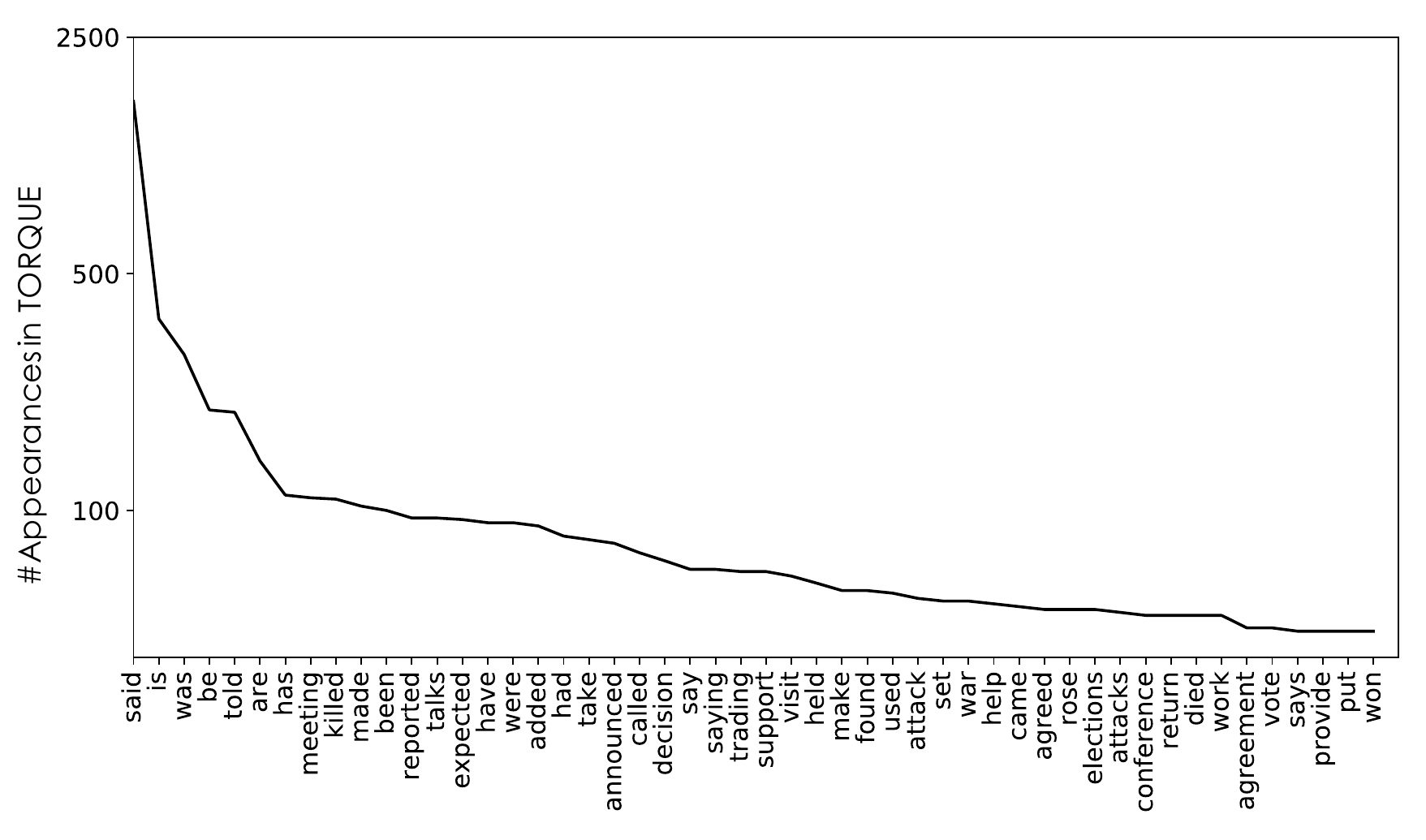}
    \caption{Fifty most common event triggers in \dataset{}. Note the y-axis is in log scale.}
    \label{fig:most common events}
\end{figure*}

Figure~\ref{fig:most common events} further shows the 50 most common events in \dataset{}. Unsurprisingly, the most common events are reporting verbs (e.g., ``say'', ``tell'', ``report'', and ``announce'') and copular verbs. Other common events such as ``meeting'', ``killed'', ``visit'', and ``war'' are also expected given that the passages of \dataset{} were taken from news articles.

\section{Question Prefix Distribution}
\label{app:sec:question}
\begin{figure*}[h!]
    \centering
    \includegraphics[width=\textwidth]{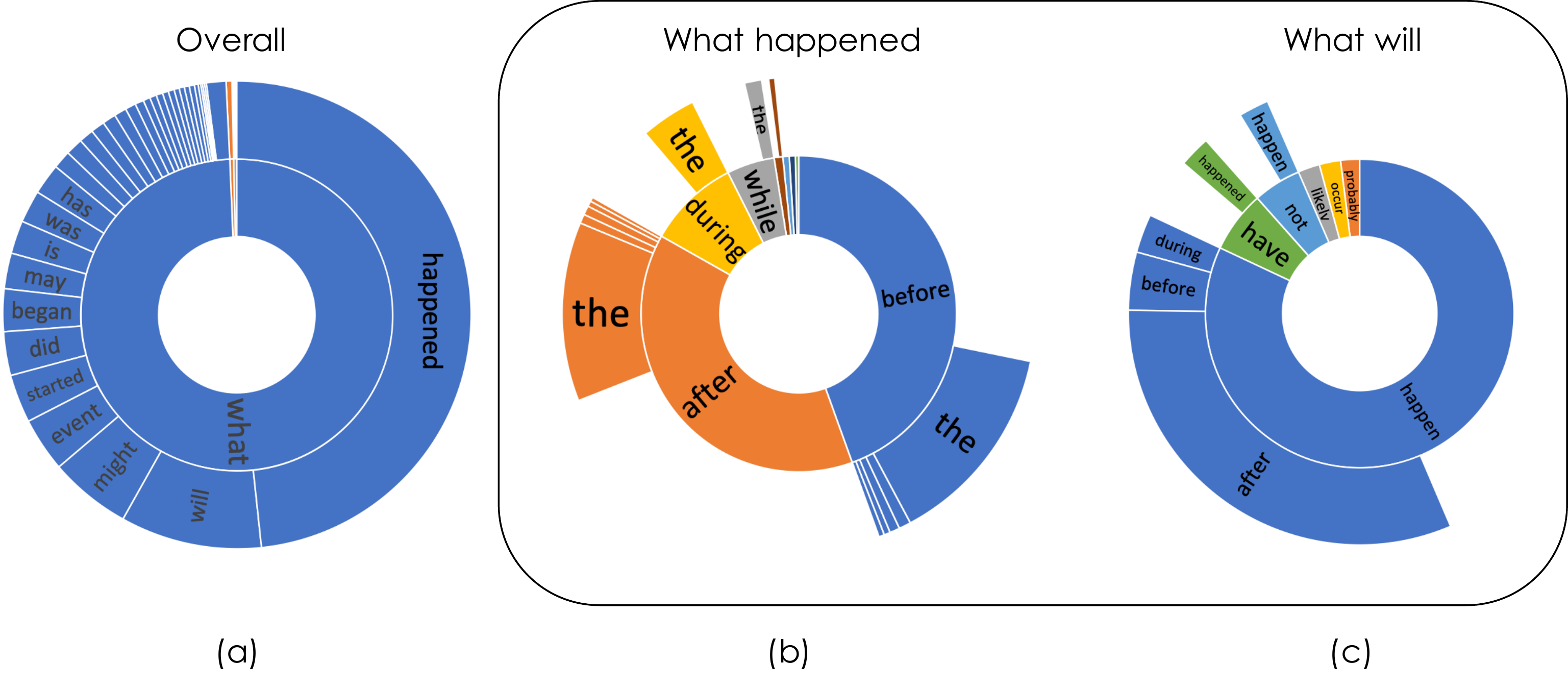}
    \caption{Prefix distribution of user-provided questions.}
    \label{fig:prefix sunburst}
\end{figure*}

Figure~\ref{fig:prefix sunburst} shows a sunburst visualization of the questions provided by crowd workers in \dataset{}, including both their original questions and their modifications.
Specifically, Fig.~\ref{fig:prefix sunburst}a shows that almost all of the questions start with ``what.'' The small portion of questions that do not start with ``what'' are cases where crowd workers switch the order of how they ask. One example of these was \question{Before making his statement to the Sunday Mirror, what did the author do?} Figure~\ref{fig:prefix sunburst}a also shows the most common following words of ``what.''

Figures~\ref{fig:prefix sunburst}b-c further show the distribution of questions starting with ``what happened'' and ``what will.'' We can see that when asking things in the past, people ask more about ``what happened before/after'' than ``what happened while/during,'' while when asking things in the future, people ask much more about ``what will happen after'' than ``what will happen before.''

\section{Answer Distribution}
\label{app:sec:answer}
\begin{figure*}[h!]
    \centering
    \includegraphics[width=\textwidth]{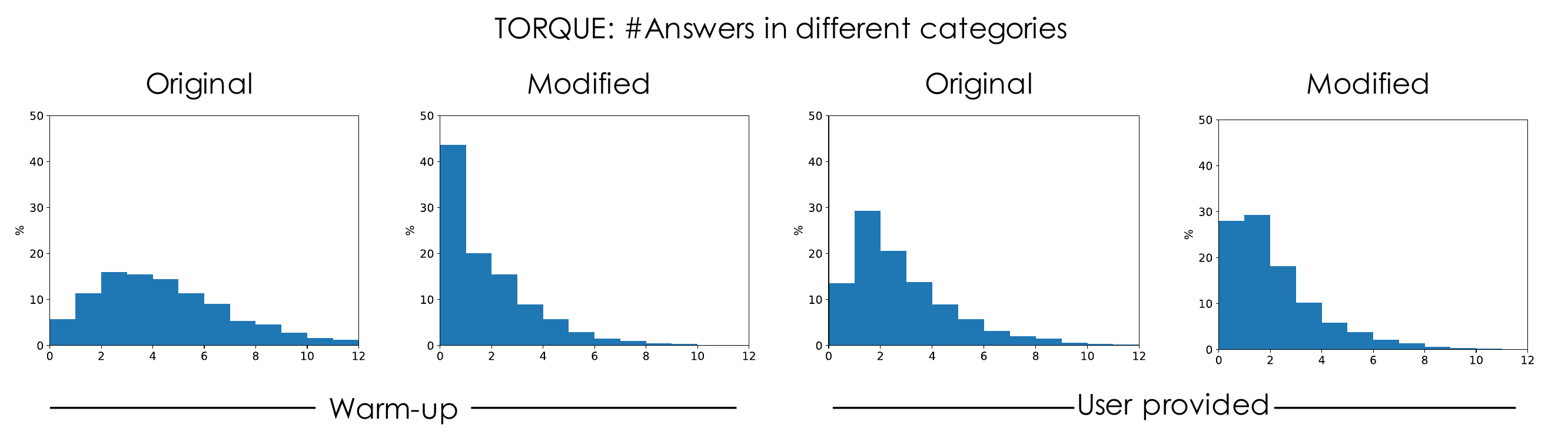}
    \caption{Distribution of the number of answers to each question.}
    \label{fig:answer distribution}
\end{figure*}

The distribution of the number of answers to each question is shown in Fig.~\ref{fig:answer distribution}, where we divide the questions into 4 categories: the original warm-up questions, the modified warm-up questions, the original user questions, and the modified user questions. Note for each passage, there are 3 warm-up questions and they were all hard-coded when crowd workers worked on them. We are treating the first one (i.e., \question{What events have already finished?}) as the original and the other two as modified (i.e., \question{What events have begun but have not finished?} and \question{What will happen in the future?}).

We can see that in both the warm-up and the user questions, ``modified'' has a larger portion of questions with no answers at all as compared to the ``original.'' This effect is very significant for warm-up questions because in news articles, most of the events were in the past. As for the user-provided questions, the percentage of no-answer questions is higher in ``modified,'' but it is not as drastic as for the warm-up question. This because we only required that the modified question should have different answers from the original one; many of those questions sill have answers after modification.

\section{Workload Distribution Among Workers}
\label{app:sec:workload}
\begin{figure*}[htbp!]
    \centering
    \includegraphics[width=.8\textwidth]{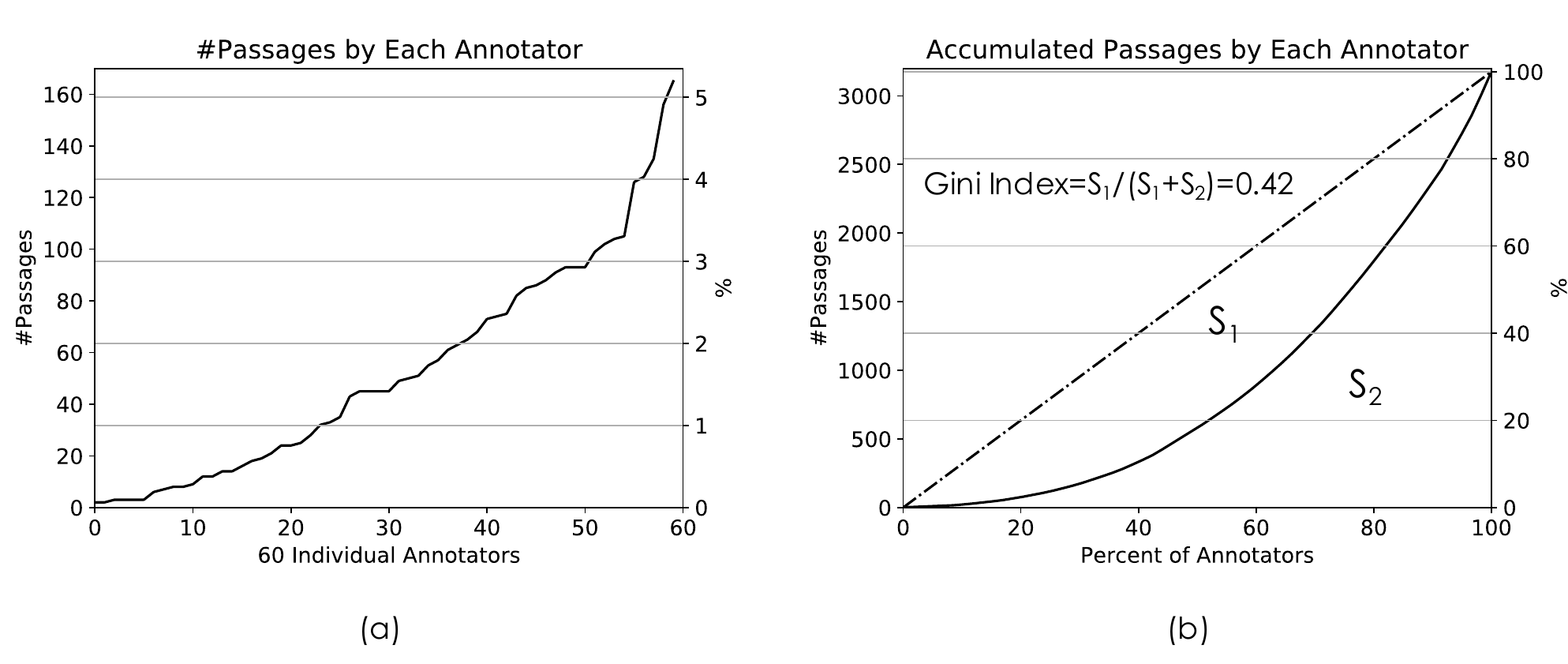}
    \caption{If every annotator provided the same number of passages (i.e., perfect equality), the curve would be the straight dashed line and the Gini Index would be 0. If one person provided all the annotations, the Gini Index is 1.}
    \label{fig:worker distribution}
\end{figure*}

As each annotator may be biased to only ask questions in a certain way, it is important to make sure that the entire dataset is not labeled by only a few annotators \citet{GevaGoBe19}. Figure~\ref{fig:worker distribution}a shows the contribution of each crowd worker to \dataset{} and we can see even the rightmost worker only provided 5\%. 
Figure~\ref{fig:worker distribution}b further adopts the notion of {\em Gini Index} to show the dispersion.\footnote{A high Gini Index here means the data were provided by a small group of workers. The Gini Index of family incomes in the United States was 0.49 in 2018 \cite{SKCM18}.} {The Gini index of \dataset{} is 0.42.} 

\section{Worker Agreement With Aggregate}
\label{app:sec:wawa}
In Sec.~\ref{sec:stats} we described the worker agreement with aggregate (WAWA) metric for measuring the inter-annotator agreement (IAA) between crowd workers of \dataset{}. This WAWA metric is explained in the figure below. It is to first get an aggregated answer set from multiple workers (we used majority vote as the aggregate function), then compare each worker with the aggregated answer set, and finally compute the micro-average across multiple workers and multiple questions.

\begin{figure*}[h!]
    \centering
    \includegraphics[width=\textwidth]{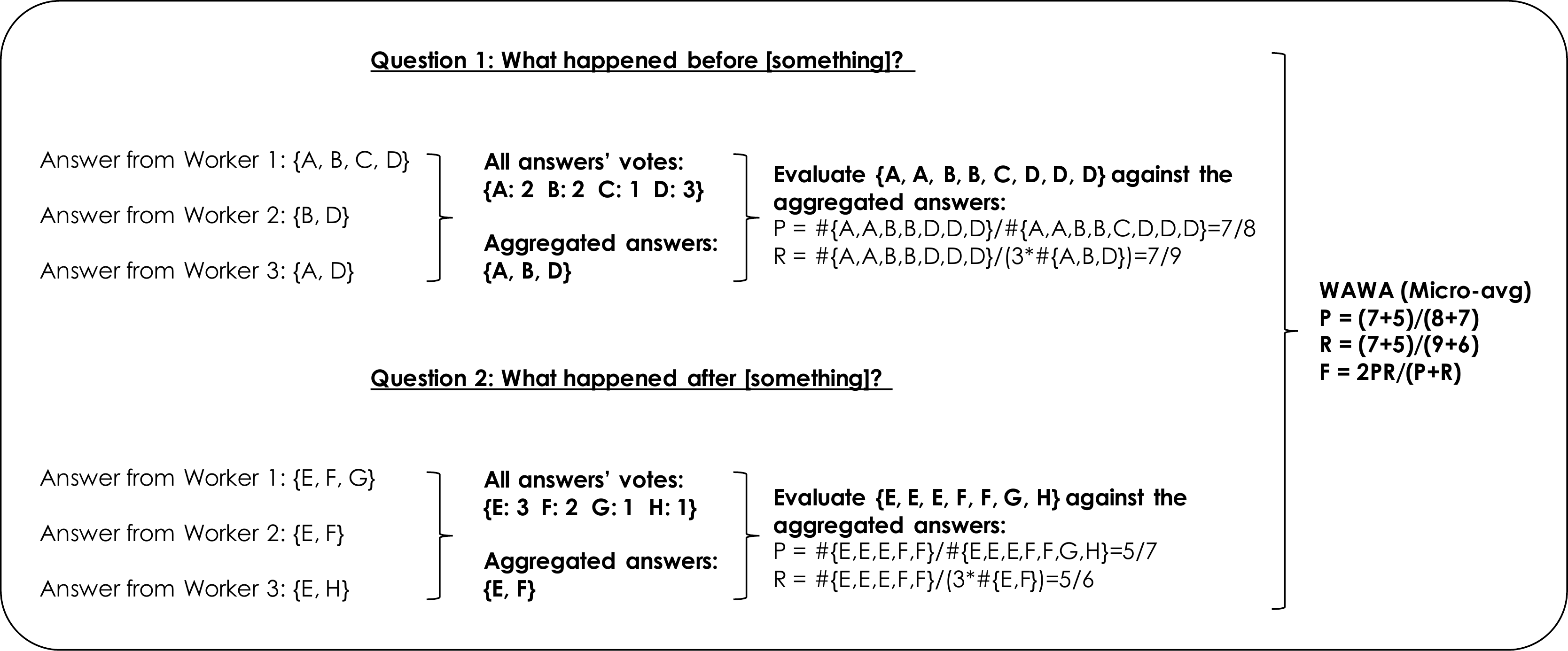}
    \caption{Explanation of the worker agreement with aggregate (WAWA) metric.}
\end{figure*}

Tables~\ref{tab:event quality} and \ref{tab:qa quality} show the quality of event annotations and question-answering annotations, respectively. In both of them, the IAA are using the WAWA metric explained above; the ``Init Annotator'' rows are a slight modification of WAWA, which means that all workers are used when aggregating those answers, but only the first annotator is compared against the aggregated answer set. Table~\ref{tab:event quality} further shows the agreement between the init annotator and an event detection model, which we have described in Sec.~\ref{sec:stats}.

\begin{table}[h!]
    \centering\small
    \begin{tabular}{ cccc } 
        \toprule
		 & P & R & F \\
		 \cmidrule(lr){2-2}\cmidrule(lr){3-3}\cmidrule(lr){4-4}
		 IAA (WAWA) & 94.3\% & 94.1\% & 94.2\%\\
		 Init Annotator & 94.9\% & 89.8\% & 92.3\%\\
		 \cmidrule(lr){1-4}
		\multicolumn{2}{c}{} & \multicolumn{2}{c}{Init Annotator}\\
		\cmidrule(lr){3-4}
		\multicolumn{2}{c}{} & Yes & No\\
		\cmidrule(lr){2-4}
		\multirow{2}{*}{Model}& Yes & 99.4\% & 82.0\%\\ & No & 64.1\% & 0.8\%\\
		\bottomrule
	\end{tabular}
    \caption{Inter-annotator agreement (IAA) of the event annotations in \dataset{}. Above: compare the aggregated event list with either all the annotators or the initial annotator. Below: how many candidates in each category were successfully added into the aggregated event list.}
    \label{tab:event quality}
\end{table}

\begin{table}[h!]
    \centering\small
    \begin{tabular}{cccc}
    \hline
         & P & R & F \\
         \cmidrule(lr){2-2}\cmidrule(lr){3-3}\cmidrule(lr){4-4}
        IAA (WAWA) & 82.3\% & 87.3\% & 84.7\%\\
        Init Annotator & 91.3\% & 82.2\% & 86.5\%\\
        \hline
    \end{tabular}
    \caption{IAA of the answer annotations in \dataset{}.}
    \label{tab:qa quality}
\end{table}

\section{Reproducibility}
\label{app:sec:reproducibility}
\begin{itemize}
    \item We ran our experiments on PyTorch 1.3.1. Pre-trained language models are implemented in the Huggingface transformers library. 
    \item A single GeForce RTX 2080 GPU was used to finetune a model. CUDA Version 10.2. The average time to run an epoch was 38 minutes for the full training section of \dataset{} using RoBERTa-large.
    \item The best performing model consist of RoBERTa-large + final MLP layer, and the number of parameters is 355M + 1024 * 64 + 64 * 2.
\end{itemize}
\end{document}